\documentclass[11pt]{article}
\usepackage{authblk}        
\usepackage{amsmath,amssymb}
\usepackage{graphicx}
\usepackage{hyperref}
\usepackage{cite}           
\usepackage{geometry}
\usepackage{tikz}
\usetikzlibrary{calc}
\usepackage{pdflscape}
\usepackage{pgfplots}
\usepackage{multirow}%
\usepackage{url}%
\usepackage{amsmath,amssymb,amsfonts}%
\usepackage{amsthm}%
\usepackage{mathrsfs}%
\usepackage[figuresright]{rotating}%
\usepackage[title]{appendix}%
\usepackage{xcolor}%
\usepackage{textcomp}%
\usepackage{manyfoot}%
\usepackage{tabularray}%
\usepackage{booktabs}%
\usepackage{algorithm}%
\usepackage{algorithmicx}%
\usepackage{algpseudocode}%
\usepackage{listings}%
\usepackage{program}%
\usepackage{textgreek}
\usepackage{caption}
\definecolor{MSwordBlue}{RGB}{23,100,188}
\definecolor{MSexcelGreen}{RGB}{75,148,78}
\definecolor{PDFmaron}{RGB}{178,34,34}
\definecolor{PPTorange}{RGB}{192,105,78}
\definecolor{Olive}{RGB}{149,156,42}
\definecolor{cadetgrey}{RGB}{57,64,69}
\definecolor{buff}{RGB}{167,141,87}

\title{Generation of Synthetic Clinical Text: A Systematic Review}

\author[1]{Basel Alshaikhdeeb}
\author[1]{Ahmed Abdelmonem Hemedan}
\author[1]{Soumyabrata Ghosh}
\author[1]{Irina Balaur}
\author[1]{Venkata Satagopam}
\affil[1]{Luxembourg Centre for Systems Biomedicine, University of Luxembourg, 6, avenue du Swing, Esch-sur-Alzette, L-4367, Luxembourg \newline
          Corresponding email: \texttt{basel.alshaikhdeeb@uni.lu}}
\date{\today}  

\begin{document}
\maketitle

\begin{abstract}
Generating clinical synthetic free-text represents an effective solution for common clinical NLP issues like sparsity and privacy. This paper aims to conduct a systematic review on generating synthetic medical unstructured free-text by formulating quantitative analysis to three research questions concerning (i) the purpose of generation, (ii) the techniques, and (iii) the evaluation methods. We searched PubMed, ScienceDirect, Web of Science, Scopus, IEEE, Google Scholar, and arXiv databases for publications associated with the generation of synthetic medical unstructured free-text. After accommodating the review on title, abstract, and full-text, we have collected 94 articles out of 1,398. A great deal of attention has been given to the generation of synthetic medical text from 2018 onwards, where the main purpose of such a generation is towards text augmentation, assistive writing, corpus building, privacy-preserving shareable medical text, annotation, and usefulness. Transformer architectures were the main predominant technique used to generate the text, especially the GPTs. On the other hand, there were four main aspects of evaluation, including similarity, privacy, structure, and utility, where utility was the most frequent method used to assess the generated synthetic medical text. Although the generated synthetic medical text demonstrated a moderate possibility to take the place of real medical documents in different downstream NLP tasks, it has proven a great asset to be augmented, complementary to the real documents, towards improving the accuracy and overcoming sparsity/undersampling issues. Yet, privacy is still a major issue behind generating synthetic medical text, where more human assessments are needed to check for the existence of any sensitive information. Despite that, advances in generating synthetic medical text will considerably accelerate the adoption of workflows and pipelines development, discarding the time-consuming legalities of data transfer.
\end{abstract}

\section{Introduction}
The extensive adoption of Electronic Medical/Health Records (EMRs/EHRs) has contributed toward generating a vast amount of clinical text associated with patient information. From clinical notes to discharge summaries, along with the lab test reports, the door has been opened widely for Natural Language Processing (NLP) tasks. However, unstructured medical free-text is facing different aspects of challenges \cite{Ford_etal_2020}. First, due to its unstructured nature, it is subject to loss because of the storing issues compared to the structural information. Unstructured medical free-text yields significant clinical details about the diagnosis, symptoms, and medications. More specifically, free-text plays a vital role in mental health notes, which highly rely on narratives that occurred between the doctor and the patient. With all that said, unstructured medical free-text is highly concerning in terms of privacy since it contains sensitive information about the patient, current disease, family history, and others. This made sharing such a medical free-text or providing public access to it highly risky since it may lead to the re-identification of individuals. On the other hand, the medical domain in general suffers from sparsity. This happens because some diseases are rare and associated with a limited number of records/documents, which poses the demand for oversampling, especially toward the minority classes/diseases \cite{NingsihETAL2022}.
The emergence of generative language models has successfully amplified the amount of unstructured free-text in which the generation is no longer limited to human manual data entry. The key characteristic behind such automated text is that it mimics the human way of formulating sentences, in which the text can be seen as synthetic, as it deviates from the original data that has been used to train the model \cite{Ren_etal_2020}. This could be a great opportunity to overcome the privacy restrictions.
Although such automatic generation of text was mainly adopted for applications other than healthcare (\textit{i.e.}, chatbot, machine translation, customer service, etc.). Yet, a great deal of attention has been given to the medical domain in the past few years. As the number of healthcare applications relying on these generative language models grows, the need to understand the utility of the generated text increases exponentially. On the other hand, there is an imperative demand to determine the capabilities among these language models, along with the assessment of their generated text based on quality, privacy, and usefulness.
This paper aims to conduct a systematic review of the synthetic medical/clinical unstructured free-text by investigating the generation purpose, techniques, and evaluation methods used to assess the generated text. The following subsections will identify the research questions, contribution, and related work.

\subsection{Research Questions (RQs)}
The main aim behind this study is to review the synthetic medical text generation. Therefore, the research questions that guided the review are as follows:
\begin{itemize}
\item \textit{RQ1: What purposes lie behind generating synthetic medical free-text? What are the languages and datasets that have been addressed?}
\item \textit{RQ2: What are the techniques used to generate synthetic medical text? What is the performance and key characteristics of these techniques?}
\item \textit{RQ3: What are the evaluation methods used to assess such a synthetic medical text? How can these methods be classified?}
\end{itemize}

\subsection{Contribution}
This study aims to conduct an extensive and narrative review to examine the synthetic text generation. This would include the purpose of such a generation, the techniques used within the generation, and the evaluation aspects that have been followed to assess such a generated text. To our knowledge, this work is the first attempt to review the generation of synthetic medical text in a comprehensive way.

\subsection{Related Work}
Examining the generation of synthetic medical data is an attractive discipline where several review efforts have been witnessed in the last five years, as depicted in Table \ref{Similar_Review}. The majority of these reviews have focused on multi-modal data generation or a specific type of data, such as image generation or tabular/structured data generation, with little attention to free-text generation in the medical domain. Although the study of  Murtaza et al. \cite{Murtaza_etal_2023} have dedicated much of their survey to medical textual data generation, they have strictly concentrated on the privacy aspect. Hence, this study will narrow down the focus to include synthetic unstructured (free-text) generation for the medical domain.

\begin{table}[h]
\begin{center}
\begin{minipage}{\textwidth}
\caption{Similar review studies}\label{Similar_Review}
\begin{tabular*}{\textwidth}{@{\extracolsep{\fill}}p{0.5cm}p{5.2cm}p{3cm}p{3cm}p{3cm}@{\extracolsep{\fill}}}
\hline%
Year & Author                                                    & Medical Synthetic Image Generation & Medical Tabular Synthetic Data Generation & Medical Multi-modal Synthetic Data Generation \\
\hline
2022 & Hernandez et al. \cite{Hernandez_etal_2022}               &                                    & \checkmark                                &                                               \\
2023 & Murtaza et al. \cite{Murtaza_etal_2023}                   &                                    &                                           & \checkmark                                    \\
2023 & Eigenschink et al. \cite{Eigenschink_etal_2023}           &                                    &                                           & \checkmark                                    \\
2024 & Sherwani \& Gopalakrishnan \cite{Sherwani_etal_2024}      & \checkmark                         &                                           &                                              \\
2024 & Ghosheh et al. \cite{Ghosheh_etal_2024}                   &                                    & \checkmark                                &                                              \\
2024 & Budu et al. \cite{Budu_etal_2024}                         &                                    & \checkmark                                &                                              \\
2024 & Pezoulas et al. \cite{Pezoulas_etal_2024}                 &                                    &                                           & \checkmark                                   \\
2024 & Kim et al. \cite{Kim_etal_2024}                           &                                    &                                           & \checkmark                                    \\
2025 & Ibrahim et al. \cite{Ibrahim_etal_2025}                   &                                    &                                           & \checkmark                                    \\
\hline
\end{tabular*}
\end{minipage}
\end{center}
\end{table}

\subsection{Paper Outline}
This paper is organized as follows: Section \ref{sec:Methods} describes in detail the process that have been followed to conduct the systematic review. In Section 3, the research questions of this study will be answered respectively. Lastly, Section 4 provides an extensive discussion where the significant findings are explained.

\section{Methods}\label{sec:Methods}
\begin{figure}[h]
\noindent
\begin{tikzpicture}


    \node[draw=cadetgrey,fill=cadetgrey, text width=3.5cm, text=white, align=center, rotate=90, inner sep=4pt] (ID) at (0.5,12.8) {\large\textbf{{Identification}}};

    \node[draw=cadetgrey,fill=cadetgrey, text width=3.2cm, align=center, text=white, rotate=90, inner sep=3pt] (SC) at (0.5,8.55) {\large\textbf{{Screening}}};

    \node[draw=cadetgrey,fill=cadetgrey, text width=3cm, align=center, text=white, rotate=90, inner sep=3pt] (EG) at (0.5,4.4) {\large\textbf{{Eligibility}}};

    \node[draw=cadetgrey,fill=cadetgrey, text width=2cm, align=center, text=white, rotate=90, inner sep=4pt] (EG) at (0.5,1) {\large\textbf{{Inclusion}}};

    \node[draw=cadetgrey,fill=cadetgrey, rounded corners, text width=7.1cm, font=\linespread{0.9}\selectfont, text=white, align=center, inner sep = 9pt] (DS) at (5,12.8) { \large{ \textbf{\underline{Database Search}}} \\[0.1in]
            \noindent\begin{tabular}{ll}
                \normalsize{PubMed = 175}            & \normalsize{arXiv = 199}          \\
                \normalsize{WoS = 326}             & \normalsize{GoogleScholar = 98}  \\
                \normalsize{ScienceDirect = 499}     & \normalsize{IEEE = 57}           \\
                \normalsize{SCOPUS = 44} \\
            \end{tabular}
       \\[0.1in] \normalsize{\textbf{(n = 1,398)}} };

    \node[draw=cadetgrey,fill=cadetgrey, rounded corners, text width=6cm, font=\linespread{0.9}\selectfont, text=white, align=center] (RBS) at (12.5,12.8) {\large{ \textbf{\underline{Remove Before Screening}}} \\[0.1in] \normalsize{Duplicated Publications} \\[0.1in] \normalsize{\textbf{(n = 675)} }};

    \node[draw=cadetgrey,fill=cadetgrey, rounded corners, text width=6cm, font=\linespread{0.9}\selectfont, text=white, align=center, inner sep = 10pt] (SCR) at (5,8.5) {\large{ \textbf{\underline{Screening}}} \\[0.2in] \normalsize{Screened Publications based on title \& abstract review} \\[0.2in] \normalsize{\textbf{(n = 723)} }};
    \node[draw=cadetgrey,fill=cadetgrey, rounded corners, text width=7cm, font=\linespread{0.9}\selectfont, text=white, align=center] (RBF) at (12.5,8.5) {\large{ \textbf{\underline{Remove Before Full-text Review}}} \\[0.1in] \small{Non-English Articles = 2 \\ No Open-access nor Institutional-access = 3 \\ Thesis = 5 \\ Review Articles = 32} \\[0.1in] \large{\textbf{(n = 42)} }};

    \node[draw=cadetgrey,fill=cadetgrey, rounded corners, text width=6cm, font=\linespread{0.9}\selectfont, text=white, align=center, inner sep = 10pt] (FTR) at (5,4.5) {\large{ \textbf{\underline{Full-text Review}}} \\[0.2in] \normalsize{Assessed Publications for Eligibility} \\[0.2in] \normalsize{\textbf{(n = 681)} }};
    \node[draw=cadetgrey,fill=cadetgrey, rounded corners, text width=7cm, font=\linespread{0.9}\selectfont, text=white, align=center] (EP) at (12.5,4.5) {\large{ \textbf{\underline{Excluded Publications}}} \\[0.1in] \tiny{Irrelevant (\textit{e.g., medical textile, next-generation sequencing, cell synthesis, clinical trials}) = 379 \\ Image Generation (\textit{e.g., ECG synthesis}) = 92 \\ Tabular Generation (\textit{e.g., computer/sensor generation}) = 52 \\ No Generation (\textit{e.g., pretrained model}) = 51 \\ Non-Medical Generation (\textit{e.g., customer support chatbot}) = 13} \\[0.1in] \normalsize{\textbf{(n = 587)} }};

    \node[draw=cadetgrey,fill=cadetgrey, rounded corners, text width=6cm, font=\linespread{0.9}\selectfont, text=white, align=center] (DC) at (5,1) {\large{ \textbf{\underline{Data Collection}}} \\[0.2in] \normalsize{\textbf{(n = 94)} }};

            \draw[cadetgrey, line width=0.3mm,-stealth] (DS) -- (RBS);
                \draw[cadetgrey, line width=0.3mm,-stealth] (DS) -- (SCR);
                    \draw[cadetgrey, line width=0.3mm,-stealth] (SCR) -- (RBF);
                        \draw[cadetgrey, line width=0.3mm,-stealth] (SCR) -- (FTR);
                            \draw[cadetgrey, line width=0.3mm,-stealth] (FTR) -- (EP);
                                \draw[cadetgrey, line width=0.3mm,-stealth] (FTR) -- (DC);

\end{tikzpicture}
\caption{Selection process flow diagram} \label{Prisma}
\end{figure}
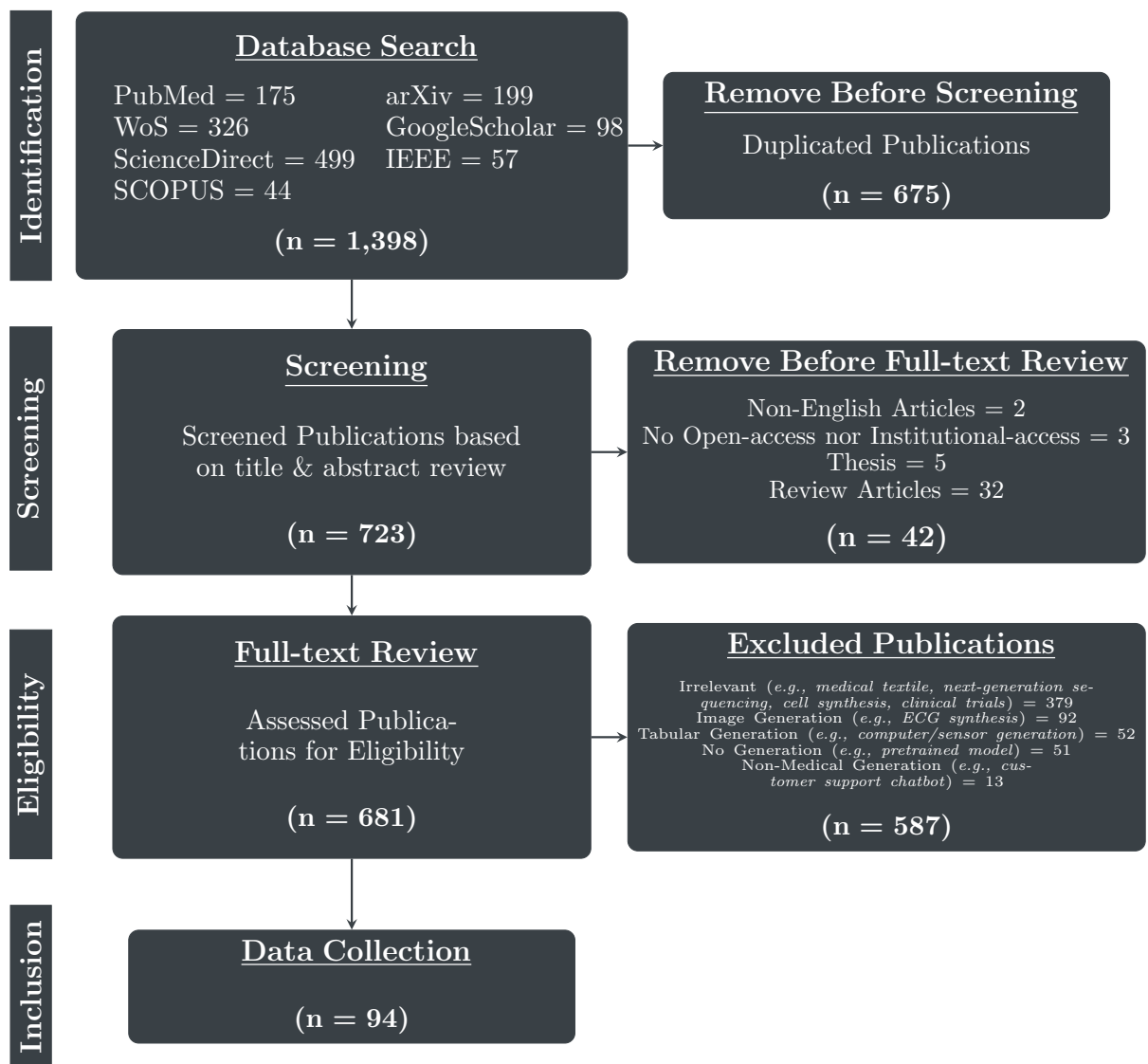

We have followed the five phases proposed by Khan et al. \cite{Khan_etal_2003} and Uman \cite{Uman_2011} to conduct the review. The first phase is represented by identifying the strategy of search. While the second phase aims at carrying out the search based on the identified strategy, where relevant articles are collected based on an inclusion criterion, and the rest are discarded based on an exclusion criterion. The third phase aims to extract the information from the selected articles through a reading and screening process. In the fourth phase, the summarization of extracted information will be conducted. Lastly, the fifth phase will aim to interpret the findings. These phases are depicted in Fig. \ref{Prisma}. The following subsections will address these methodological phases.

\subsection{Search Strategy}
To formulate the strategy of search, it is necessary first to identify the search engines. In fact, we have selected seven search engines for this systematic review, including PubMed\footnote{https://pubmed.ncbi.nlm.nih.gov} (RRID: SCR\textunderscore004846), ScienceDirect\footnote{https://www.sciencedirect.com} (RRID: SCR\textunderscore027174), Web of Science\footnote{http://webofscience.com} (RRID: SCR\textunderscore022629), Scopus\footnote{https://www.scopus.com} (RRID: SCR\textunderscore022613), IEEE\footnote{https://ieeexplore.ieee.org} (RRID: SCR\textunderscore008314), Google Scholar\footnote{https://scholar.google.com} (RRID: SCR\textunderscore008878), and arXiv\footnote{https://arxiv.org} (RRID: SCR\textunderscore005488). These search engines would provide a wide coverage of general scientific publications.
On the other hand, we have set three search limits, including the search term, date, and type of publication. First, for the search terms, since our review focuses on synthetic text generation for the medical domain, the roots of the four terms \textit{synthetic}, \textit{medical}, \textit{text}, and \textit{generation} have been selected to be searched in the title as follows:
\\\textit{(generat* OR augment*) AND (synthe* OR pseudo* OR artifici*) AND (medic* OR clinic* OR health*) AND (text* OR record* OR note*)}
\\Then, much more rich key phrases with further synonyms have been used to search the title, abstract, keywords, and/or topic fields as follows:
\\\textit{("synthetic text" OR "synthetic free-text" OR "synthetic unstructured text" OR "synthetic natural language") AND ("health records" OR "medical records" OR "clinical notes") AND ("generative model" OR "data augmentation" OR "re-use")}
\\For the date of publications, we believe that the breakthrough of transformer architecture in “Attention is all you need” \cite{Vaswani_etal_2017} (i.e., from 2017 onwards) has significantly contributed toward the rise of text generative models. Yet, we preferred to set the publication date from 2015 until the time of bibliographic search, which is August 2024, in order to include earlier efforts. Lastly, for the publication type, we have limited our search to peer-reviewed journals and conference papers written in the English language. We have also included some non-peer-reviewed articles from arXiv in order to get some new insights.

\subsection{Inclusion Criteria}
In terms of the publication type, we have mainly focused on peer-reviewed journals and conference articles. Yet, we have added some non-peer-reviewed articles from arXiv since it contains the latest trends. In addition, we have included publications written in English and accessed through either institutional or open-access. We have focused mainly on articles rather than theses, dissertations, and posters. On the other hand, since the aim of this review is to focus on generating synthetic medical free-text, the selected articles must meet three requirements. First, the article must include a mechanism for generating medical unstructured free-text. Second, the article must show a purpose for such a generation. Third, the generated synthetic text should be inferred from EMR/EHR datatypes (\textit{i.e.}, clinical notes, discharge summaries, patient records, lab reports, etc.).

\subsection{Exclusion Criteria}
The articles that do not meet the inclusion criteria will be discarded. This would include i) synthetic non-medical text generation, ii) synthetic medical non-text (structured, tabular, images, etc.) generation, iii) no generation method, iv) theses, dissertations, or posters, v) non-English written publications, and finally, vi) neither institutional-accessed nor open-access publications.

\subsection{Data Extraction}
Once the data has been collected according to the inclusion and exclusion criteria, the data extraction will take place. The extracted data will reflect RQ1, in which the purposes of generating synthetic medical text are determined, along with the languages and datasets that have been addressed. Meanwhile, it will also include information in response to RQ2 where the techniques of generation are discussed, accompanied by the key features of each technique. Lastly, in response to RQ3, the evaluation methods used to assess the generated synthetic medical text will be extracted and categorized.

\subsection{Selected Articles}
As depicted in Fig. \ref{Prisma}, the search begins with popular databases, which led to a total of 1,398 articles. After applying the criteria, a total of 94 relevant articles have been selected (See \textbf{Appendix \ref{sec:SelectedArticles}} for the complete table list).

\section{Results}\label{sec:Results}
\subsection{RQ1: What purposes lie behind generating synthetic medical free-text? What are the languages and datasets that have been addressed?}
\begin{figure}
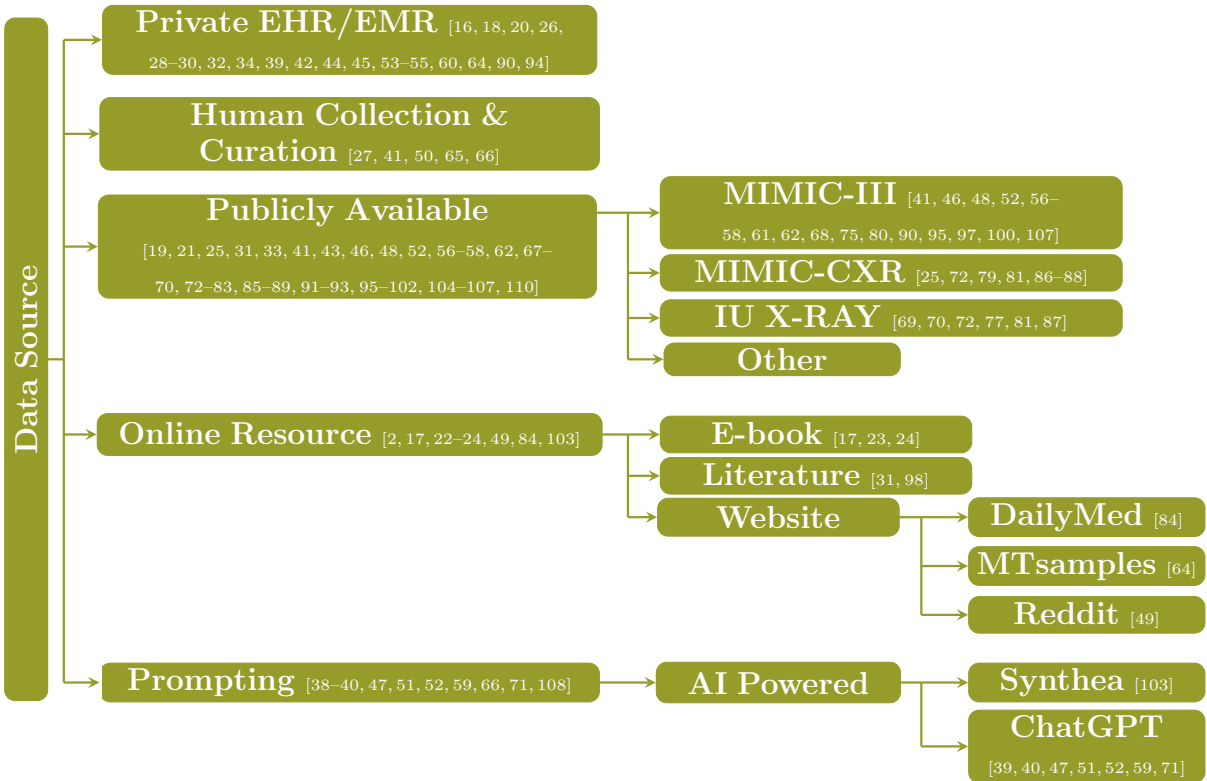

\noindent
%
\caption{Datasets} \label{Datasets}
\end{figure}

The synthetic medical text must have been generated to serve a particular goal. In essence, the literature showed six main aims, including \textit{Privacy-Preserving}, \textit{Augmentation}, \textit{Usefulness}, \textit{Assistive Writing}, \textit{Annotation}, and \textit{Corpus Building}, as shown in Fig. \ref{Purposes} (See Appendix \ref{sec:Purposes} for more elaboration).
\\While the majority of the studies have examined the generation using the English language, there are some efforts that have been made to serve Chinese \cite{PengETAL2019,QUETAL2023,Guan_etal_2018,XiaETAL2022,Guan_etal_2019,HuangETAL2022}, German \cite{BorchertETAL2020,Modersohn_etal_2022,Lohr_etal_2018}, Japanese \cite{NishinoETAL2020,KagawaETAL2021,Igarashi_Nihei_2022}, Norwegian \cite{LundETAL2024,Rama_etal_2018,Brekke_etal_2021}, French \cite{HiebelETAL2023}, Dutch \cite{Libbi_etal_2021}, Arabic \cite{ALMutairiETAL2024}, Indonesian \cite{NingsihETAL2022}, and Bulgarian \cite{Velichkov_etal_2020}.
\\In terms of the data that has been used to train the generation model, there were five sources depicted in the literature, including \textit{Private EHR/EMR}, \textit{Manual Collection/Curation}, \textit{Online Resources}, \textit{Prompting}, and \textit{Publicly Available Sources} as depicted in Fig. \ref{Datasets}. Private EHR/EMR refers to the studies that have harnessed discharge summaries or medical notes from hospitals. Whereas, the manual creation refers to the studies that collected and curated clinical text manually. On the other hand, online resources refer to the utilization of E-books, literature, or medical websites (\textit{e.g.}, DailyMed\footnote{https://dailymed.nlm.nih.gov} , Mtsamples\footnote{https://mtsamples.com} , Reddit\footnote{https://www.reddit.com}). Prompting refers to the process of utilizing AI-powered tools (\textit{e.g.}, Synthea\footnote{https://synthea.mitre.org} , ChatGPT\footnote{https://openai.com/index/chatgpt}) where some studies have obtained their initial data using such tools. Lastly, publicly available sources refer to the gold standard/benchmark datasets, including MIMIC-III \cite{Johnson_etal_2016}, MIMIC-CXR \cite{Johnson_etal_2020}, and IUX-RAY \cite{Demner-Fushman_etal_2015}.

\subsection{RQ2: What are the techniques used to generate synthetic medical text? What is the performance and key characteristics of these techniques?}
In fact, there are four general categories of techniques that have been used for generating synthetic medical text in the literature, including \textit{Manual}, \textit{Text Processing}, \textit{Knowledge Source}, and \textit{Neural Network} models, as shown in Fig. \ref{GenTechs} (See Appendix \ref{sec:Techniques} for further elaboration).

\begin{figure}
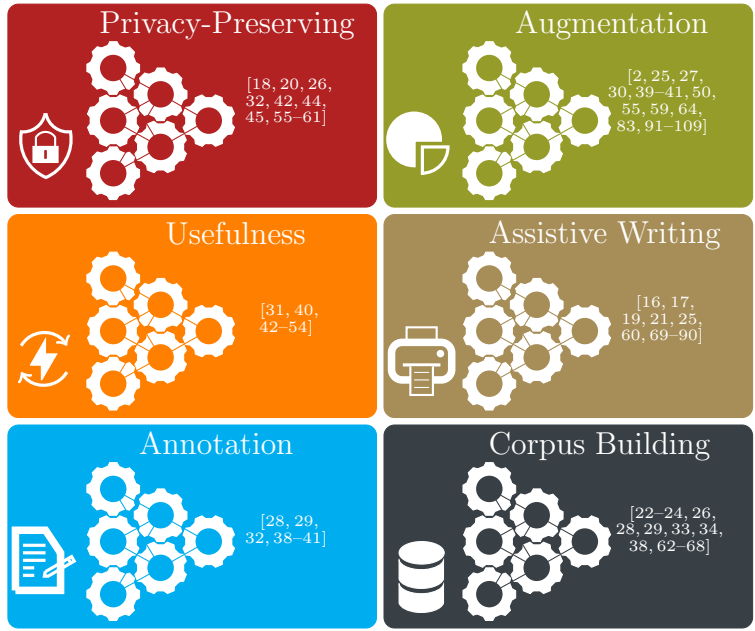

\noindent
\centering

\caption{Generation Purposes} \label{Purposes}
\end{figure}

\subsection{RQ3: What are the evaluation methods used to assess such a synthetic medical text? How can these methods be classified?}
\begin{landscape}
    \begin{figure}
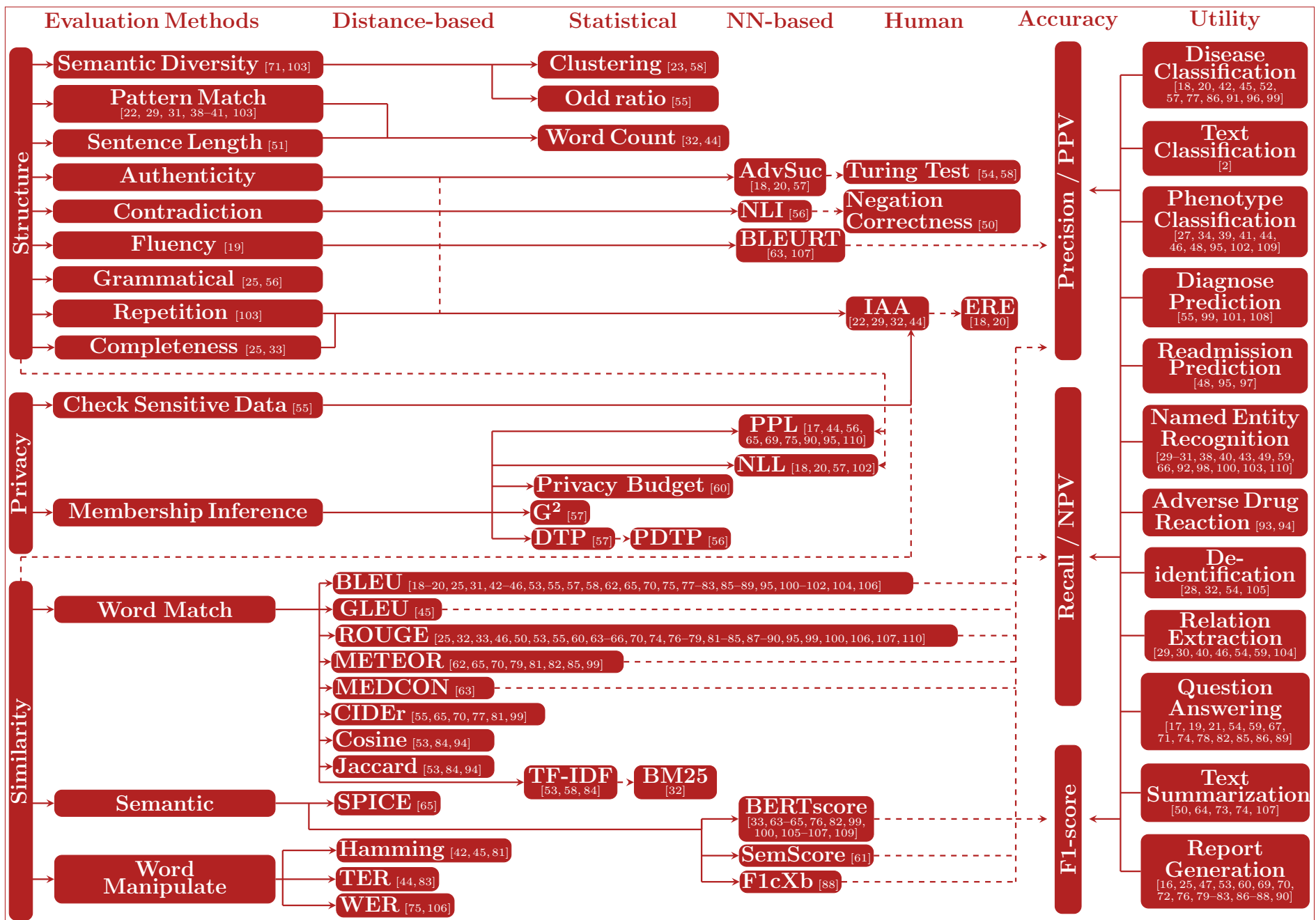

%
    \caption{Evaluation Methods} \label{EvalTechs}
\end{figure}
\end{landscape}

In order to evaluate the generated synthetic text, four aspects have been addressed within the literature, including \textit{Structure}, \textit{Privacy}, \textit{Similarity}, and \textit{Utility}, as shown in Fig. \ref{EvalTechs}. Each evaluation aspect has its own automatic (\textit{i.e}., distance-based, statistical, and Neural Network-based) and manual (\textit{i.e}., Human) metrics (See Appendix \ref{sec:Evaluation} for further elaboration).

\begin{landscape}
    \begin{figure}
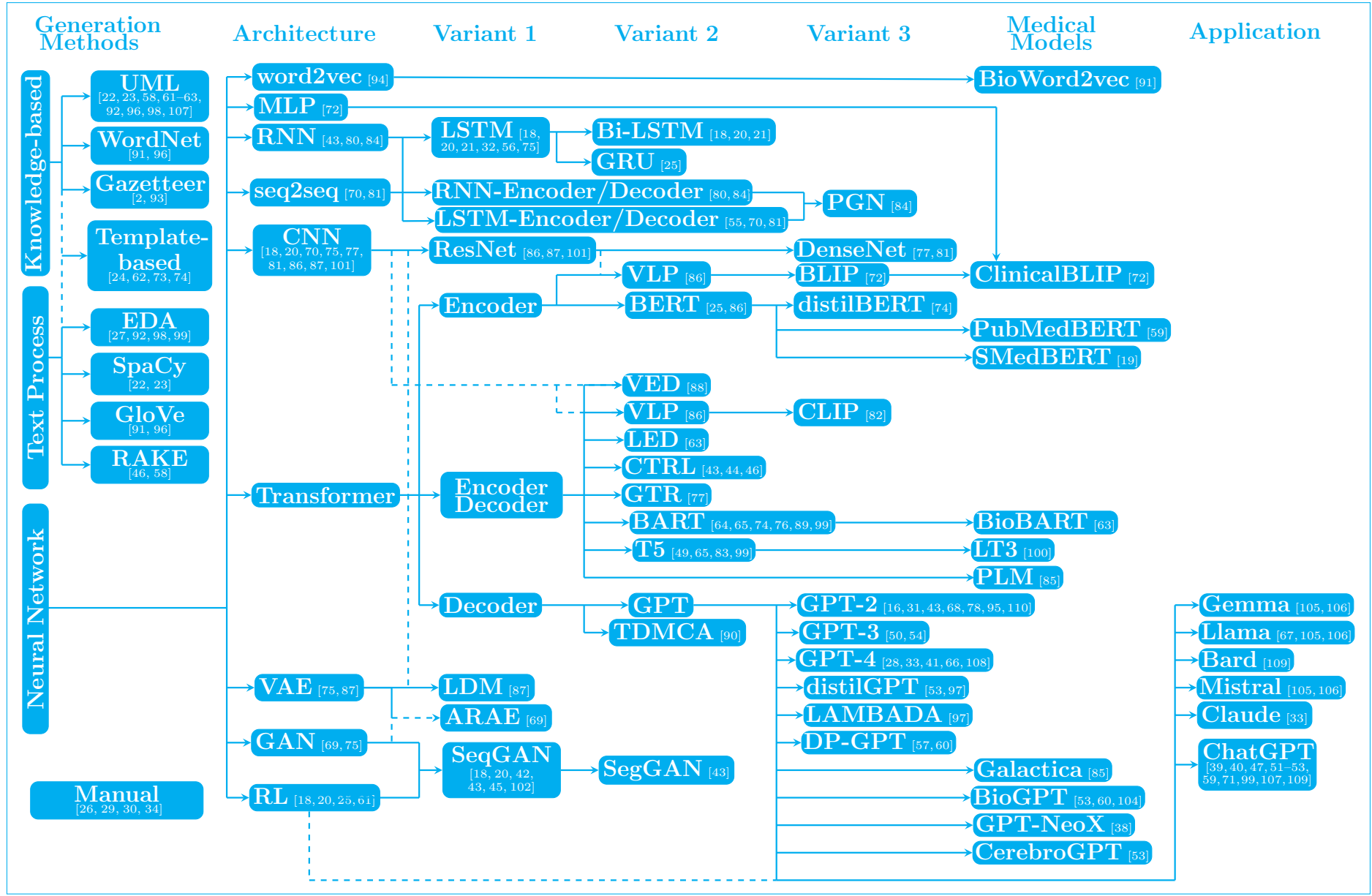


        \caption{Generation Techniques} \label{GenTechs}
\end{figure}
\end{landscape}

\section{Discussion}
Fig. \ref{Stat} shows the general statistics of the selected articles and the following subsections will provide an extensive discussion on the advantages and challenges of generating medical synthetic text.

\subsection{Benefits of Synthetic Medical Text Generation}
As depicted in the literature, generating synthetic medical text will bring valuable contributions. For example, the automated generation of clinical reports along with the autocompletion will save hundreds of hours spent by clinicians and physicians manually filling clinical notes and generating reports. On the other hand, the generation of synthetic medical text has potential in the field of annotation and labelling, where it can save time on manual curation of large-scale text corpora by generating labelled text. In addition, the synthetic generation of medical text showed better privacy capabilities than using only the de-identification method. Lastly, when the generated synthetic medical text was assessed in terms of usefulness, it showed the possibility to take the place of original text in some cases; meanwhile, it demonstrated magnificent capability to augment the original data in almost all cases. This can be a proof of concept for overcoming the problem of undersampling when handling imbalanced medical datasets. Lastly, the synthetic generation would significantly accelerate the pace of research, where the longest time is often consumed on agreements between parties or even compliance with regulated policies.

\subsection{Privacy Challenges}
Although generating synthetic medical text can overcome the shortcomings of the de-identification method, such a method is still important and should be applied prior to the generation itself. It has been noticed from one of the studies that generating synthetic text using raw data (\textit{i.e.}, without de-identification) led to generating realistic identities. Even though researchers concluded that this might be a natural protection feature against re-identification. Yet, it is still much safer to accommodate the de-identification before the generation.
\\Even with synthetic generation over de-identified text, privacy still represents the main challenge behind generating synthetic clinical/medical text due to the possibility of re-identification and membership inference threats caused by multiple factors. For instance, the least frequent and triggering phrases, such as (\textit{the accident was widely reported in the press}) can be easily used to trace the identity of an individual. In addition, copying exact and longer sequences from the original text during the generation would increase the possibility of re-identification. This is due to the existence of unique treatment, medications, and tests that could be involved in these sequences. In other cases, synthetic text was acting as a continuation of the original text, which would reveal sensitive information.
\\However, the most crucial privacy concern is memorizing information about certain individuals within the training model that generated the synthetic text. Hence, the day when we can witness the release of publicly available and privacy-preserving patient data is not yet to come.

\subsection{Structural Challenges}
Several structural shortcomings have been depicted in the literature during the generation of synthetic medical text. Misspelling, ambiguous abbreviations, and repeated terms were products of the generation. In addition, grammatical and syntactic incorrectness have been witnessed during the generation, such as replacing pronouns (\textit{e.g.}, he instead of she), incorrect medical phrases (\textit{e.g.}, Hepatitis C deficiency), and incorrect orders of discharge summaries, which usually contain admission details, medical history, treatment, and medications. Less coherence has been depicted as well, albeit surprisingly it did not harm the usefulness of the synthetic text since it was meant to train machine learning models and not for teaching humans. Lastly, the text diversity within the generated synthetic text has a significant impact on downstream tasks. It has been noticed that generating synthetic medical text using a small corpus within the training has led to less generalized text with less diversity and therefore, poor quality.

\subsection{Generation Methods}
While manual generation of synthetic text might be seen as accurate, it is highly expensive, cumbersome, and time-consuming. Using an external knowledge source, by contrast, can facilitate data augmentation but not fully synthetic text generation. Within all the techniques, the transformer models were the most promising ones. However, it has been noticed that the generation of text through traditional or so-called vanilla transformer was not viable, especially if the training data was relatively small. This is because the traditional transformer has difficulty with longer sequence modelling tasks due to the limitation of 512 tokens as a maximum processing capacity. This is also applied on BERT, where it meant not to generate text but rather serving a particular task by fine-tuning such as extracting an answer to a question or predicting a predefined label of text.
\\The most adequate models were represented by the GPTs due to the massive amount of data that has been used to train them along with the ability to modify its determinism through the hyperparameters, specifically the so-called temperature. It is worth mentioning that no conclusive evidence has been observed regarding the benefits of medical-specific pretrained models towards generating text. The reason behind such a limitation is due to the lack of understanding of colloquial tone within the text by such models.  On the other hand, GANs showed some potential. The key characteristic behind GAN lies in the discriminator architecture, which parallelly trained with the generator architecture. In this case, the generation of text will be tuned based on the ability of the discriminator to distinguish the real from the synthetic. While some studies demonstrated the outperformance of GPTs over GAN, SeqGAN (which boosts the GAN with RL) and conditional GAN are still promising. The concept of conditional generation has shown a great performance where the generation of text is guided or conditioned by contextual information. Supplementing contextual information to GPTs would be significantly promising.

\subsection{Evaluation Challenges}
While the similarity metrics are important tools to measure the closeness between the generated text and the real one, they may contradict with the privacy. In fact, the efforts toward boosting the similarity would lead to revealing unique medical components associated with individuals, which implies the need for a trade-off mechanism. On the other hand, there is also a drawback behind using the automatic privacy metrics, which cannot give a full guarantee of privacy-preserving of the generated synthetic text. The need for human assessment during privacy evaluation is imperative, where a set of healthcare or privacy professionals could carefully check the generated text. This is also applied to the text structure assessment, where postprocessing can be performed by experts to correct invalid syntax. Lastly, multiple utilities need to be tested in order to investigate the usefulness of the generated synthetic text. In other words, some tasks or utilities would be negatively impacted by the random generation of text. For example, considering a generated synthetic medical text that mixes terms such as hypertension and hypotension within the same patient record would lead to producing inadequate data instances for a downstream task of phenotype classification.

\begin{landscape}
\begin{figure}[htbp]
\centering

\begin{minipage}[t]{0.48\linewidth}
\centering
\begin{tikzpicture}
\begin{axis}[
    width=\linewidth,
    height=3cm,
    ymin=0, ymax=40,
    bar width=0.5cm,
    symbolic x coords={2015,2018,2019,2020,2021,2022,2023,2024,2025},
    xtick=data,
    nodes near coords,
    nodes near coords style={text=MSwordBlue, font=\scriptsize},
    title={\textbf{(a) Year}},
    ylabel={\tiny{\textbf{\# of Articles}}},
    xticklabel style={font=\tiny, align=center,yshift=-1.6em},
    yticklabel style={rotate=0}
]
\addplot+[ybar, mark=none, draw=MSwordBlue, fill=MSwordBlue]
    coordinates {(2015,1) (2018,8) (2019,7) (2020,7) (2021,13) (2022,14) (2023,15) (2024,24) (2025,4)};
\end{axis}
\end{tikzpicture}
\end{minipage}
\hfill
\begin{minipage}[t]{0.48\linewidth}
\centering
\begin{tikzpicture}
\begin{axis}[
    width=\linewidth,
    height=3cm,
    bar width=0.5cm,
    symbolic x coords={Arabic,Bulgarian,Dutch,French,Indonesian,Japanese,Norwegian,German,Chinese,English},
    xtick=data,
    title={\textbf{(b) Language}},
    nodes near coords,
    nodes near coords style={text=MSexcelGreen, font=\scriptsize},
    ylabel={\tiny{\textbf{\# of Articles}}},
    ymin=0, ymax=90,
    xticklabel style={font=\tiny, rotate=40, anchor=east, align=center}
]
\addplot+[ybar, mark=none, draw=MSexcelGreen,fill=MSexcelGreen] coordinates {(Arabic,1) (Bulgarian,1) (Dutch,1) (French,1) (Indonesian,1) (Japanese,3) (Norwegian,3) (German,4) (Chinese,6) (English,75)};
\end{axis}
\end{tikzpicture}
\end{minipage}

\vspace{0.1cm} 

\begin{minipage}[t]{0.48\linewidth}
\centering
\begin{tikzpicture}
\begin{axis}[
    width=\linewidth, 
    height=3cm, 
    ymin=0,
    ymax=100,
    bar width=0.5cm,
    symbolic x coords={Usefulness,Annotation,Privacy,Corpus Building,Assistive Writing,Augmentation},
	xticklabels={{Usefulness},{Annotation},{Privacy},{Corpus\\Building},{Assistive\\Writing},{Augmentation}},
    xtick=data,
    nodes near coords,
    nodes near coords style={text=PPTorange, font=\scriptsize},
    title={\textbf{(c) Purpose}},
    ymin=0, ymax=40,
    ylabel={\tiny{\textbf{\# of Articles}}},
	xticklabel style={font=\tiny,align=center},
    yticklabel style={rotate=0}
]
\addplot+[ybar, mark=none, draw=PPTorange,fill=PPTorange] coordinates {(Usefulness,5) (Annotation,6) (Privacy,13) (Corpus Building,13) (Assistive Writing,27) (Augmentation,30)};
\end{axis}
\end{tikzpicture}

\end{minipage}
\hfill
\begin{minipage}[t]{0.48\linewidth}
\centering
        	\begin{tikzpicture}
\begin{axis}[
    width=\linewidth, 
    height=3cm,
    bar width=0.5cm,
    symbolic x coords={Online source,Prompting,Private EHR/EMR,IU X-RAY,MIMIC-CXR,MIMIC-III,Other},
    xticklabels={{Online\\source},{Prompting},{Private\\EHR/EMR},{IU\\X-RAY},{MIMIC-\\CXR},{MIMIC-\\III},{Other}},
    xtick=data,
    nodes near coords,
    nodes near coords style={text=Olive, font=\scriptsize},
    title={\textbf{(g) Data Source}},
    ymin=0, ymax=40,
    ylabel={\tiny{\textbf{\# of Articles}}},
	xticklabel style={font=\tiny,align=center},
    yticklabel style={rotate=0}
]
\addplot+[ybar, mark=none, draw=Olive,fill=Olive] coordinates {(Online source,8) (Prompting,10) (Private EHR/EMR,19) (IU X-RAY,4) (MIMIC-CXR,7) (MIMIC-III,17) (Other,27)};
\end{axis}

\end{tikzpicture}
\end{minipage}

\vspace{0.1cm} 

\begin{minipage}[t]{0.48\linewidth}
\centering
    	\begin{tikzpicture}
\begin{axis}[
    width=\linewidth, 
    height=3cm,
    bar width=0.2cm,
    symbolic x coords={UML,WordNet,Gazetteer,Template-based,EDA,SpaCy,GloVe,RAKE,word2vec,MLP,RNN,LSTM,CNN,Seq2Seq,Transformers,BERT,GPT,VAE,GAN,RL,Manual},
    xtick=data,
    nodes near coords,
    nodes near coords style={text=cyan, font=\scriptsize},
    title={\textbf{(d) Generation Technique}},
    ymin=0, ymax=60,
    ylabel={\tiny{\textbf{\# of Articles}}},
    xticklabel style={rotate=45, anchor=east},
	xticklabel style={font=\tiny,align=center},
    yticklabel style={rotate=0}
]
\addplot+[ybar, mark=none, draw=cyan,fill=cyan] coordinates {(UML,10) (WordNet,2) (Gazetteer,2) (Template-based,4) (EDA,4) (SpaCy,2) (GloVe,2) (RAKE,2) (word2vec,2) (MLP,1) (RNN,2) (LSTM,9) (CNN,9) (Seq2Seq,6) (Transformers,28) (BERT,5) (GPT,46) (VAE,3) (GAN,8) (RL,8) (Manual,4)};
\end{axis}
\draw[decorate,decoration={brace,amplitude=6pt},thick,yshift=-70pt]
  (1.8,0) -- (0.5,0) node[midway,below=5pt] {\textit{\small{Knolwedge-base}}};

\draw[decorate,decoration={brace,amplitude=6pt},thick,yshift=-30pt]
  (3.5,0) -- (2,0) node[right=30pt,below=5pt] {\textit{Text Processing}};

\draw[decorate,decoration={brace,amplitude=6pt},thick,yshift=-70pt]
  (8,0) -- (3.7,0) node[midway,below=5pt] {\textit{Neural Network}};

            \draw[decorate,decoration={brace,amplitude=6pt},thick,yshift=-30pt]
  (9.5,0) -- (8.5,0) node[midway,below=5pt] {\textit{Manual}};
\end{tikzpicture}
\end{minipage}
\hfill
\begin{minipage}[t]{0.48\linewidth}
\centering
        	\begin{tikzpicture}
\begin{axis}[
    width=\linewidth, 
    height=3cm,
    bar width=0.2cm,
    symbolic x coords={SPICE,SemScore,F1cXb,BM25,GLEU,MedCon,WER,TER,Cosine,Hamming Distance,TFIDF,Jaccard,CIDEr,METEOR,BERTscore,BLEU,ROUGE,Privacy Budget,Check Sensitive Information,DTP,NLL,PPL,Odd-ratio,Clustering,NLI,AdvSuc,BLEURT,Pattern Match,Human Assessment,Classification Accuracy, Utility},
    xtick=data,
    nodes near coords,
    nodes near coords style={text=PDFmaron, font=\scriptsize},
    title={\textbf{(e) Evaluation Methods}},
    ymin=0, ymax=70,
    ylabel={\tiny{\textbf{\# of Articles}}},
    xticklabel style={rotate=45, anchor=east},
	xticklabel style={font=\tiny,align=center},
    yticklabel style={rotate=0}
]
\addplot+[ybar, mark=none, draw=PDFmaron,fill=PDFmaron] coordinates {(SPICE,1) (SemScore,1) (F1cXb,1) (BM25,1) (GLEU,1) (MedCon,1) (WER,2) (TER,2) (Cosine,2) (Hamming Distance,3) (TFIDF,3) (Jaccard,4) (CIDEr,6) (METEOR,9) (BERTscore,12) (BLEU,36) (ROUGE,47) (Privacy Budget,1) (Check Sensitive Information,2) (DTP,2) (NLL,4) (PPL,9) (Odd-ratio,1) (Clustering,2) (NLI,2) (AdvSuc,3) (BLEURT,3) (Pattern Match,8) (Human Assessment,41) (Classification Accuracy,34) (Utility,59)};
\end{axis}

            \draw[decorate,decoration={brace,amplitude=6pt},thick,yshift=-70pt]
  (5,0) -- (0.5,0) node[midway,below=5pt] {\textit{Similarity}};

\draw[decorate,decoration={brace,amplitude=6pt},thick,yshift=-30pt]
  (6,0) -- (4.9,0) node[midway,below=5pt] {\textit{Privacy}};

\draw[decorate,decoration={brace,amplitude=6pt},thick,yshift=-70pt]
  (8,0) -- (6.1,0) node[midway,below=5pt] {\textit{Structure}};

                \draw[decorate,decoration={brace,amplitude=6pt},thick,yshift=-30pt]
  (9,0) -- (8,0) node[midway,below=5pt] {\textit{Utility}};

\end{tikzpicture}
\end{minipage}

\vspace{0.1cm} 

\begin{minipage}[t]{0.48\linewidth}
\centering
    \begin{tikzpicture}
\begin{axis}[
    width=\linewidth, 
    height=3cm, 
    ymin=0,
    ymax=100,
    bar width=0.5cm,
    symbolic x coords={Text Classification,Adverse Drug Reaction Extraction,Readmission Prediction,De-Identification,Diagnose Prediction,Text Summarisation,Relation Extraction,Phenotype Classification,Disease Classification,Question Answering,Named Entity Recognition,Report Generation},
	xticklabels={{Text\\Classification},{Adverse Drug\\Reaction Extraction},{Readmission\\Prediction},{De-Identification},{Diagnose\\Prediction},{Text\\Summarisation},{Relation\\Extraction},{Phenotype\\Classification},{Disease\\Classification},{Question\\Answering},{Named Entity\\Recognition},{Report\\Generation}},
    xtick=data,
    nodes near coords,
    nodes near coords style={text=PDFmaron, font=\scriptsize},
    title={\textbf{(f) Utility}},
    ymin=0, ymax=30,
    ylabel={\tiny{\textbf{\# of Articles}}},
    xticklabel style={rotate=45, anchor=east},
	xticklabel style={font=\tiny,align=center},
    yticklabel style={rotate=0}
]
\addplot+[ybar, mark=none, draw=PDFmaron,fill=PDFmaron] coordinates {(Text Classification,1) (Adverse Drug Reaction Extraction,2) (Readmission Prediction,3) (De-Identification,4) (Diagnose Prediction,5) (Text Summarisation,6) (Relation Extraction,7) (Phenotype Classification,10) (Disease Classification,11) (Question Answering,13) (Named Entity Recognition,14) (Report Generation,18)};
\end{axis}
\end{tikzpicture}
\end{minipage}
\hfill
\begin{minipage}[t]{0.48\linewidth}
\centering
    \begin{tikzpicture}
\begin{axis}[
    width=\linewidth,
    height=3cm,
    bar width=0.5cm,
    symbolic x coords={COVID news,Endoscopy Reports,Medical Consultation,Clinical Transcripts,PICO,Clinical Practice Guidelines,Patient Laboratory Test,Biological Concepts \& Relations,Drug \& Medication,History of Present Illness,Radiology,Doctor-Patient Conversations,Discharge Summaries },
	xticklabels={{COVID\\news},{Endoscopy\\Reports},{Medical\\Consultation},{Clinical\\Transcripts},{PICO},{Clinical Practice\\Guidelines},{Patient\\Laboratory Test},{Biological Concepts\\ \& Relations},{Drug \& \\Medication},{History of\\Present Illness},{Radiology},{Doctor-Patient\\Conversations},{Discharge\\Summaries}},
    xtick=data,
    title={\textbf{(h) Clinical Domain}},
    nodes near coords,
    nodes near coords style={text=MSexcelGreen, font=\scriptsize},
    ylabel={\tiny{\textbf{\# of Articles}}},
    ymin=0, ymax=40,
    xticklabel style={rotate=45, anchor=east},
	xticklabel style={font=\tiny,align=center},
    yticklabel style={rotate=0}
]
\addplot+[ybar, mark=none, draw=MSexcelGreen,fill=MSexcelGreen] coordinates {(COVID news,1) (Endoscopy Reports,1) (Medical Consultation,2) (Clinical Transcripts,2) (PICO,3) (Clinical Practice Guidelines,3) (Patient Laboratory Test,5) (Biological Concepts \& Relations,6) (Drug \& Medication,6) (History of Present Illness,10) (Radiology,13) (Doctor-Patient Conversations,14) (Discharge Summaries,27) };
\end{axis}
\end{tikzpicture}
\end{minipage}

\caption{General Statistics}\label{Stat}
\end{figure}
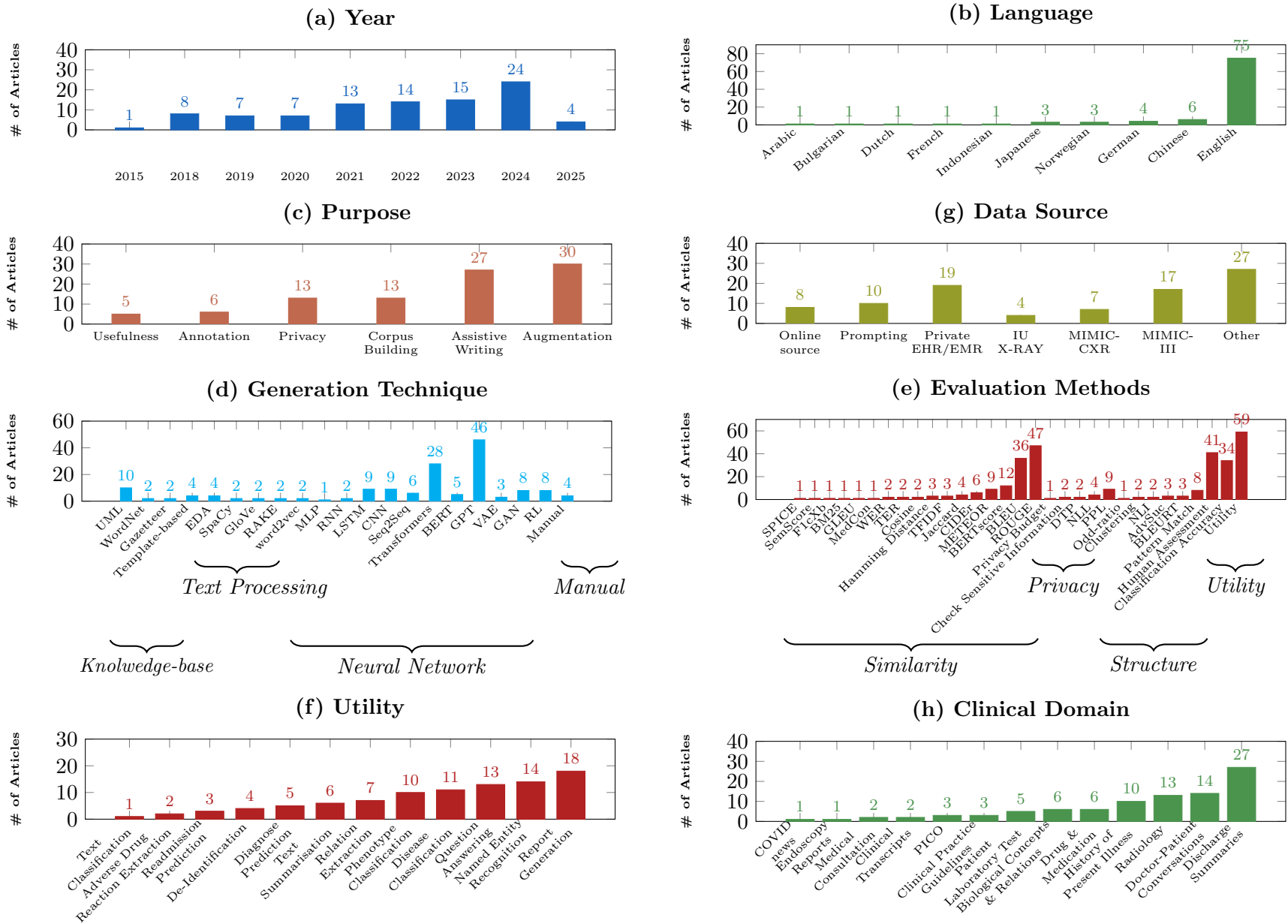
\end{landscape}

\subsection{Limitations}
The main limitation behind the methodical review in this study is the possible selection bias inherent to the search methods used. The most remarkable one is the polysemous of the term synthetic which has different context meaning the clinical trials and may retrieve large number of relevant publications to this review. For that, we have tried to use different synonyms such during the search such as artificial and pseudo. On the other hand, the use of root terms such as \textit{generat\textsuperscript{*}} during the search, has not worked properly when using some databases and retrieve zero result. For that, we have used the multiple possibilities of the word itself such as \textit{generate}, \textit{generation}, \textit{generative}, and \textit{generator}. Further derivational inflections might not be examined during our search. Another methodical limitation lies in the automated medical text generation in general. There are definitely much more publications relevant to that subject yet, since our study focused on synthetic which means making changes on the text itself, we have tried to adhere to the studies that attempted to make changes on text during the automation.
\\An organizational limitation of the review given by this study lies behind the lack of proper comparison of the performance among generation techniques due to the absence of unified evaluation metrics. Some studies have concentrated on privacy while others focused on how similar the synthetic text is compared to the original. In addition, the lack of publicly available datasets of EMRs/EHRs due to privacy concerns has also contributed towards this limitation. Another limitation was depicted by the lack of illustration for the non-neural-network-based methods compared to the illustration give to the neural network architectures. The reason is that we believe that neural language models have potential in the future and will gain much more attention than the traditional techniques.

\section{Conclusion}
The role of unstructured free-text in medical domain has never been greater to analyse useful information within EMRs/EHRs such as phenotype prediction, disease classification or even extracting word-level artifacts including medications, lab tests, symptoms, and others. With the great challenge of sharing unstructured medical text due to privacy, the synthetically generation of such a text could be an alternative solution. Not only to overcome privacy but also towards tackling undersampling problems where medical unstructured datasets are predominantly suffering from unbalancing. This paper has provided a comprehensive taxonomy of the synthetic medical text generation where purpose, techniques, and evaluation used within the generation have been extensively discussed. We argue that medical synthetic text generation is expected to play significant roles in different future downstream analysis. For future direction, conducting an empirical comparison of the generation techniques would provide more comprehensive understanding of the pros and cons.

\section{Acknowledgement}
This work was supported by the CHIST-ERA grant CHIST-ERA-22-ORD-02, by the Luxembourg National Research Fund (FNR, INTER/CHIST23/17882238/FAIRClinical).

\pagebreak

\appendix

\section{Appendix A: Generation Purposes}
\label{sec:Purposes}
\subsection{Privacy-Preserving}
In the medical field, privacy plays a vital role, in which preserving individual information is an essential task. Western cultures like Anglo-American and European countries consider the individual’s right to own his/her personal data; thus, regulations have been formed to protect this ownership. In this sense, privacy violation would simply refer to the disclosure of individual information through multiple threats, including Re-identification, Membership Inference, and Attribute Disclosure. Re-identification (also known as de-anonymization) refers to the ability to reveal the identity of an individual from his/her medical data \cite{PengETAL2023}. Membership inference is an attack that is intended to test whether an individual was a part of data collection using a combination of features of that individual \cite{Lee_2018_Natural}. This might not be limited to a dataset but also a pre-trained machine learning model. Attribute Disclosure refers to revealing sensitive attributes of an individual.
\\A rule has been regulated by the USA Health Insurance Portability and Accountability Act (HIPAA) along with General Data Protection Regulation (GDPR) by the European Union for publishing/sharing of patient electronic health records, including clinical text. Such a regulation permits the publication/sharing of such records within strict conditions of de-identification. De-identification is the process of eliminating personal identifiers such as name, location, phone number, dates (\textit{e.g.}, birthdate, admission date, etc.), hospital, and others, which can be referred to as Personally Identifiable Information (PII) or Protected Health Information (PHI).
\\The public view toward structural medical data has widely been examined, where different studies showed participants were more comfortable sharing and publishing their medical information through statistical figures \cite{Aitken_etal_2016}. Yet, most of the public view investigation studies regarding sharing medical data did not explicitly differentiate between whether the data is structured or unstructured \cite{Stockdale_etal_2019}. In this regard, Ford et al. \cite{Ford_etal_2020} have presented a study that specified the public view toward sharing unstructured medical free-text within a citizen jury study in the UK. Results of such a study showed greater agreement in favour of sharing medical free-text for research purposes, but under strict conditions where a proper de-identification process would take place.
\\Yet, there are cases where de-identification itself would seem insufficient due to the potential re-identification of an individual through its distinct combination of clinical events or unique clinical data elements. Therefore, synthetically generated medical text could be an alternative solution.

\subsection{Augmentation}
NLP approaches are considered data-hungry or data-greedy, which indicates the high demand for more samples in order to gain better performance. In addition, sharing language resources such as corpora, ontologies, dictionaries, annotations, and lexical resources is the backbone of the NLP infrastructure toward the progress of its tasks. In that essence, privacy would obstruct valuable contributions that could be achieved through sharing NLP resources. On the other hand, dealing with medical text data , for tasks such as diagnosis/disease classification usually require handling the problem of undersampling. This happens because some diseases are rare and associated with a limited number of records/documents, which poses the demand for oversampling, especially toward the minority classes/diseases \cite{NingsihETAL2022}. Hence, generating synthetic medical text can facilitate overcoming this problem by increasing the size of instances associated with minority disease classes.

\subsection{Annotation}
Some NLP tasks like NER, ADR, RL, or even the automatic de-identification require ground truth curation that is usually conducted by experts. Suppose a biomedical relation, such as a protein-protein interaction, exists within a piece of text. The need to identify such a relation requires an expert in the biology domain, where he/she would curate the exact sentence that articulated such a relation, along with the entities themselves (\textit{i.e.}, Protein A and Protein B). This is to enable learning through the annotated examples by the machine learning models. This is considered to be the backbone of the training process and would significantly impact the quality of the model. Yet, labelling or curating such artefacts by the experts is seen as a cumbersome and time-consuming task, especially if the size of the text is huge. Hence, generating already annotated medical text would save much time where the expert/curator would only need to review the generated tags rather than creating them from scratch.

\subsection{Corpus Building}
Similar to the annotation purpose, corpus building is the process of preparing textual data examples for training the machine learning models. The corpus might serve tasks like NER and RL, but also can be extended to serve much detailed tasks like text summarization and question answering. This definitely would require text sample acquisition, cleaning, and curation, which can be discarded by the generation of synthetic text. 

\subsection{Assistive Writing}
The medical staff, including medical receptionists, medical secretaries, medical administrative assistants, or even clinicians, are often required to write reports regarding a particular patient. Such a task might seem time-consuming and involve errors. Therefore, generating clinical reports through synthetic medical text might save plenty of time where the medical staff is only asked to review such a text rather than creating it from scratch. This would considerably enhance the healthcare quality through accelerating the service, which leads to a shorter wait time of wait by patients.

\subsection{Usefulness}
Generating new knowledge is predominantly associated with testing the utility of such knowledge. Therefore, this purpose aims to view the benefits of generated medical synthetic text towards NLP tasks. Note that it might overlap with other purposes such as the annotation, augmentation, or even privacy-preserving.
\pagebreak

\section{Appendix B: Generation Techniques}
\label{sec:Techniques}

\subsection{Manual}
Manual techniques refer to the use of human experts in the medical domain to formulate, curate, and review clinical text \cite{Velichkov_etal_2020,Rama_etal_2018,KagawaETAL2021,Brekke_etal_2021}.  The common methodologies used in these techniques are crowdsourcing and human-in-loop, where the task of generating or curating text is divided into sub-tasks that integrate a group of people to get their interventions and feedback.

\subsection{Text Processing}
Text processing techniques refer to the adoption of traditional semi-automatic approaches to recreate the original medical text. This kind of approach includes Easy Data Augmentation \textit{EDA} \cite{Igarashi_Nihei_2022,LatifETAL2024,Issifu_Ganiz_2021,Kang_etal_2020}, \textit{SpaCy}\footnote{https://allenai.github.io/scispacy/} \cite{BorchertETAL2020,Modersohn_etal_2022}, Global word-to-word co-occurrence Vectors \textit{GloVe} \cite{Abdollahi_etal_2020,Abdollahi_etal_2021}, Template-based \cite{NguyenETAL2023,GoldsteinAyelet2015,Lohr_etal_2018,Begoli_etal_2018}, and Rapid Automatic Keyword Extraction \textit{RAKE}\footnote{https://github.com/csurfer/rake-nltk}) \cite{Zhou_etal_2022_DataSifterText,Wang_etal_2019_artificial}. EDA relies on four main steps: Synonym Replacement \textit{SR}, Random Insertion \textit{RI}, Random Swap \textit{RS}, and Random Deletion \textit{RD}. SR aims to replace non-stopword terms with their semantic synonyms through the use of an external knowledge source. Whereas, RI refers to the process of adding a synonym for a random non-stopword term in a random position. RS refers to the process of switching two randomly chosen terms. Finally, RD aims at eliminating a random term. In EDA, only the textual input \textit{x} is being altered while the target \textit{y} will be preserved. Whereas, SpaCy  is a Python package that serves NLP tasks such as tokenization, part-of-speech tagging, and entity recognition. Similarly, RAKE is an algorithm for extracting keywords and key phrases based on frequent occurrences. GloVe factorized the global co-occurrence of terms within a large text corpus (\textit{i.e}., English Wikipedia) in order to generate embedding vectors. Lastly, template-based refers to the approaches that aim at identifying predefined patterns so-called patterns such as dates, named entities, and events.

\subsection{Knowledge-based}
knowledge source techniques depend on external dictionaries or ontologies in the medical domain, which include \textit{Gazetteer} \cite{Tao_etal_2019,NingsihETAL2022}, \textit{WordNet} \cite{Abdollahi_etal_2020,Abdollahi_etal_2021}, and the Unified Medical Language \textit{UML} \cite{YimETAL2023,Zhou_etal_2022_DataSifterText,MeoniETAL2024,BorchertETAL2020,Modersohn_etal_2022,SchlegelETAL2023,Issifu_Ganiz_2021,Abdollahi_etal_2021,Begoli_etal_2018,Kang_etal_2020}. WordNet, which is an open domain English lexicon that encodes the semantic relations among terms such as synonyms (\textit{i.e.}, similar), antonyms (\textit{i.e.}, opposite), hypernyms (\textit{i.e.}, broad term or is-a), and hyponyms (\textit{i.e.}, item in a group or part-of). UML is a medical ontology that contains semantic medical annotations such as disease, syndrome, symptom, pharmacologic substances, and other qualitative and functional concepts associated with medical and biomedical interactions.

\subsection{Neural Network Architectures}
\begin{figure}
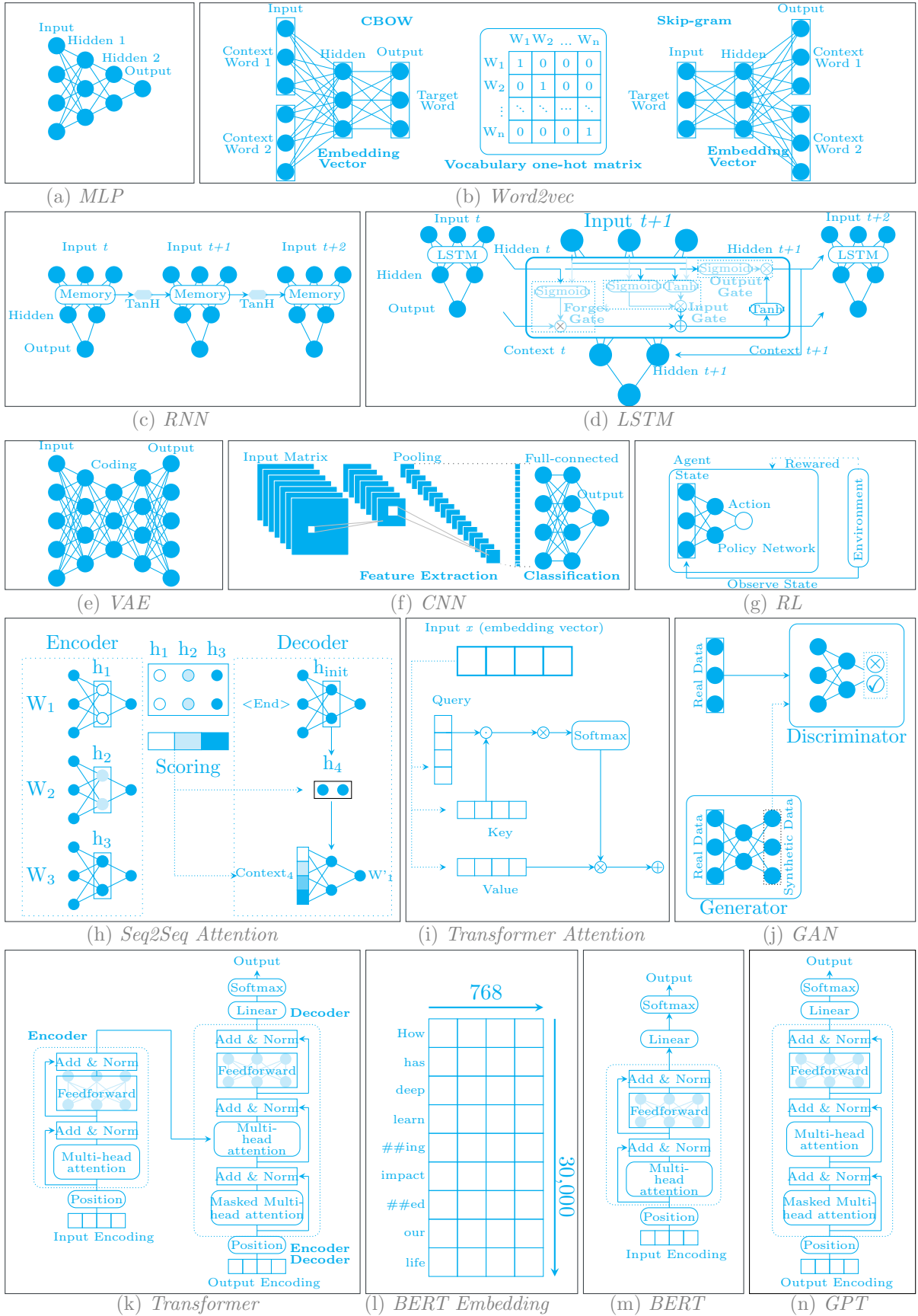


    \caption{Neural Network Architectures} \label{NNarch}
\end{figure}

The traditional Neural Network \textit{NN}, or Multilayer Perceptron \textit{MLP} (also known as vanilla) consists of three main layers: input, hidden, and output (See Fig. \ref{NNarch}(a)). The input layer resembles the features of an input data \textit{x} while the output layer represents the desired target \textit{y}, which corresponds to the class label. The hidden layer represents the encoding of the relationships among the features. Note that the hidden layers could be ranged from single to multiple layers (\textit{i.e.}, two or three) \cite{Ramchoun_etal_2016}. The encoding of the features within the hidden layer is performed through the product of generating random weights and the input features’ values, with a constant value added known as bias. This will be repeated from layer-to-layer (\textit{i.e.}, input-to-hidden, hidden-to-hidden, and hidden-to-output). The final result will be fed into an activation function in order to add non-linearity to the architecture and considered as the predicted target. There are a variety of activation functions such as Hyperbolic Tangent \textit{TanH}, \textit{Sigmoid}, \textit{Softmax}, and others. A comparison will be conducted between the actual and predicted target to calculate the error rate. Hence, the training or learning mechanism of NN will be acquired through a concept known as Backpropagation, which aims to proceed backward and update the weights’ values. This will be repeated until the error rate becomes marginal (\textit{i.e.}, the difference between actual and predicted target is minimal). At that moment, the final weight values will be stored and used for future prediction of unseen data (\textit{i.e.}, testing data). Although MLP has not been frequently used within the clinical text generation yet, it has been integrated with much more sophisticated architectures that will be depicted in the following subsections.

\subsubsection{Word2vec}
This version of NN is a fully connected feedforward architecture that aims to handle text data. Considering the One-hot encoding, which aims to represent categorical labels within a sparse matrix (\textit{i.e.}, a matrix full of zero vectors with only a single \textit{1} element corresponding to the match). Textual data can be handled within the one-hot encoding through a sparse matrix of vocabulary containing the unique terms. Hence, each one-hot vector of each term will be fed into the \textit{word2vec} architecture in order to predict the one-hot vector of another term \cite{Goldberg_etal_2014}. A similar training of generating random weights and updating them through the backpropagation will take place. Once the difference between the actual and predicted target is minimized, the hidden neuron values will be stored as Word Embeddings (WEs). The resulting embedding vector will contain contextual and relational information in which the relationship between the two words \textit{male} and \textit{king} would be similar to the one between the two words \textit{female} and \textit{queen}.
\\In fact, Word2Vec has two different architectures; \textit{Skip-gram} and Continuous Bag of Words \textit{CBOW} (See Fig. \ref{NNarch}(b)). Skip-gram aims to process a one-hot vector of a target term, attempting to predict its contextual terms. This is useful to identify the word’s semantics. Whereas CBOW aims to process the one-hot vectors of the contextual terms, attempting to predict a target term, this is useful to identify the syntax around the word. The resulting embedding is supposed to yield similarity in syntax and semantics. This means that similar words should have closer values within their embedding vectors.
\\However, word2vec architecture is highly impacted by the training text size, in which it requires a large size of text within the training in order to have accurate embedding. Therefore, there were multiple research efforts aiming to train word2vec on large text and then provide the embedding as pretrained vectors that can be used for further downstream tasks (\textit{e.g.}, classification, prediction, or clustering). \textit{BioWord2Vec} is an effort that examined the problem of domain-specific embedding by training the word2vec architecture on a large biomedical/medical text from PubMed and generated pretrained vectors \cite{Zhang_etal_2019_BioWord2Vec}.
\\Yet, word2vec had a remarkable problem known as Out-of-Vocabulary \textit{OOV}. This problem refers to the cases where an individual word would have no embedding simply due to its absence during the training.

\subsubsection{Recurrent Neural Network (RNN)}
In an attempt to handle time series or sequential data, \textit{RNN} has been introduced. RNN aims at preserving some information from the input at a time \textit{t}, attempting to pass it into future input (\textit{i.e.}, input \textit{t\textsubscript{i}}). In fact, RNN has the typical structure of layers as the traditional NN yet it has an additional layer called memory or context layer (See Fig. \ref{NNarch}(c)). Such a memory will pass such contextual information through an activation function (usually TanH) \cite{Zaremba_etal_2014}. In the context of NLP, RNN has been used in a so-called \textit{CharRNN} architecture to generate character-level embedding rather than word-level embedding in word2vec. Since the number of characters is fixed, this mechanism has shown better alleviation of the OOV problem. Li et al. \cite{Li_etal_2021} have used CharRNN architecture to generate synthetic clinical text for the purpose of privacy-preserving.

\subsubsection{Long Short Term Memory (LSTM)}
In some scenarios, the long-term contextual information could be needed to predict the current label. For example, consider the scenario of predicting the following word for a large text corpus where the prediction would take place at the end while the required contextual information lies at the very beginning of the corpus. In this case, it is highly difficult for the RNN architecture to keep long-term contextual information. In addition, long sequences are the main cause of a problem called \textit{vanishing gradient}, which refers to the periodic reduction of neuron values resulting from the continuous product of derivatives through many hidden layers, leading to values closer to zero. Here comes the \textit{LSTM}, which is an extension of RNN to overcome the problem of preserving long-term contextual information. LSTM has a similar structure to the RNN except that its memory is much sophisticated \cite{Graves_2012}. The memory cell of LSTM contains three main components called gates: \textit{forget gate}, \textit{input gate}, and \textit{output gate} (See Fig. \ref{NNarch}(d)). The forget gate contains a sigmoid function that determines whether to forget or keep information, where it produces a value between 0 (\textit{i.e.}, completely forgetting) to 1 (\textit{i.e.}, completely preserving). On the other hand, the input gate contains two activation functions, including sigmoid and TanH. Sigmoid will determine which value within the context to be updated, while TanH will replace the old (\textit{i.e.}, need to be updated) information from the context by a vector of the new candidates.
\\Lastly, the output gate aims at preparing what to output, where it contains the same activation functions. Yet, it applies the sigmoid first to determine what part of the context would contribute to the current output, then it applies the TanH to concatenate such a part to the output. 
\\LSTM has different variants; one of these variants is the Bi-directional LSTM \textit{Bi-LSTM}, which will apply the same concept in forward and backward manners sequentially. This mechanism has proven high efficacy at handling natural languages where it took advantage of word ordering and gained much more contextual information \cite{Arisoy_etal_2015}.
\\Another variant of LSTM is the Gated Recurrent Unit \textit{GRU}, which simplified the memory cell of the LSTM by combining the forget and input gates \cite{Chung_etal_2014}.
\\However, both RNN and LSTM have encountered a remarkable drawback during generating text. Such a problem is known as \textit{exposure bias}, where the model is trying to predict the next token using an incorrect previous token. The incorrectness of the previous token comes from the fact that the model has not seen the true previous token during the training; thus, replacing it with one of its generated tokens. Such a problem will be amplified with longer text, causing inaccurate results.

\subsubsection{Sequence-to-Sequence (Seq2Seq)}
Although LSTM has considerably alleviated the problem of vanishing gradient, such a problem remains as the number of words is growing. In other words, the longer the text given to the LSTM, the more loss in long-term contextual information is going to happen. On the other hand, both RNN and LSTM showed a very slow performance during the training when handling longer sequences. For this purpose, Luong et al. \cite{Luong_etal_2015} presented an architecture known as \textit{Seq2Seq} consisting of encoder and decoder layers for machine translation where RNN has been used within the two layers (See Fig. \ref{NNarch}(h)). Seq2Seq aims to make the learning much more concentrated on specific parts of the input sequence rather than focusing on too much information. Unlike the traditional RNN, where the accumulated hidden state is passed through the time intervals, Seq2Seq aims to let the encoder pass all the individual hidden states of all time intervals to the decoder. Consequently, every vector of the hidden states will be scored in order to identify the relationships among the input sequence; in other words, determining which hidden states yield the most needed information for the current time interval. Then, the decoder will perform a Softmax on each hidden state and multiply it by the score. The resulting vector will then be fed as an input to predict the next token. Seq2Seq has been depicted in the literature where, in some cases, the encoder/decoder was either RNN, LSTM, or even GRU. A modified variant of Seq2Seq is the Pointer Generator Network \textit{PGN} \cite{see2017pointsummarizationpointergeneratornetworks}, which has been intended to solve the problem of OOV by employing the pointer mechanism, which aims at copying text portions from the input within the generation.

\subsubsection{Variational Auto-Encoder (VAE)}
Another special case of encoder-decoder architecture is VAE which aims at processing an input \textit{x} and attempts to predict the same \textit{x} input data at the output \cite{Kingma_Welling_2013}. It consists of three internal layers including \textit{encoding}, \textit{coding}, and \textit{decoding} (See Fig. \ref{NNarch}(e)). Encoding layer aims at learning either a lower or higher dimensionality of the input data. Coding will then learn a much lower dimensionality of the data where the continuous latent variables are being preserved. After that, decoding will learn to rebuild a higher dimensionality from the latent variables within the coding. Lastly, the output will resemble the actual/exact dimension of the input data. VAE has been widely used for dimensionality reduction and feature extraction due to its mechanism of learning lower dimensionality of the data \cite{Hou_2017_Deep}. 

\subsubsection{Convolutional Neural Network (CNN)}
Another deep neural network architecture which has been widely used for image processing, pattern recognition, and character-to-character language models is the \textit{CNN} \cite{OShea_etal_2015}. Its structure deviated from the traditional NN by introducing a new component known as feature extraction along with the traditional component of fully connected layers for classification (See Fig. \ref{NNarch}(f)). The feature extraction component aims at performing a filtering process through three mechanisms: \textit{convolutional layers}, \textit{pooling}, and \textit{strides}. Convolutional layers are intended to produce filters with specific two-dimensional size (2 × 2 or 3 × 3). Hence, the strides will determine how many desired steps to move the filters. Lastly, the pooling aims to conduct averaging among the filters' values where maximization or minimization could be performed. Then, the resulting matrix will be flattened and fed into a fully-connected layer to perform the classification. An integrated architecture of CNN and VAE known as Latent Diffusion Model \textit{LDM} \cite{rombach2022highresolutionimagesynthesislatent} has been depicted in the literature to generate text aligned with medical images, taking advantage of latent variables of the images rather than their pixels.

\subsubsection{Transformer}
Vaswani et al.\cite{Vaswani_etal_2017} have leveraged the concept of attention towards introducing a new architecture known as \textit{Transformer}. Such an architecture consists of an \textit{encoder} and a \textit{decoder} that apply the concept of attention. The key difference is that the attention in transformer took advantage of parallelization in order to process multiple words simultaneously. In comparison to Bi-directional LSTM that addresses the order of words in two directions (\textit{i.e.}, forward and backward) sequentially, or the Seq2Seq that pays attention to different words sequentially, the attention in the transformer can process multiple parts of the sequence input in a parallel mode. This alleviates the problem of slow training in LSTM and RNN, which gives an efficient processing by the transformer. The transformer is composed of two components, including encoder and decoder, each of which takes an embedding vector; the first takes input and the second takes its corresponding output (\textit{e.g.}, sentence and its translation). Then, a positional embedding is added to the embedding vector. Such a positional encoding aims at adding the information of a token’s position for better understanding of the word’s context. Hence, the attention will take place followed by a feedforward NN that will output the attention weight. Note that there are two encoders; one on the left side and the other on the right side, called encoder-decoder (See Fig \ref{NNarch}(k)). The encoder-decoder attention performs an additional task of masking, where the next token is masked to let the model predict such a word. The output of both encoders will then be passed to the decoder that is intended to perform another attention mechanism to map the relations among the two encoders’ outputs, followed by a feedforward NN. Lastly, a linear transformation and a Softmax will be applied towards the desired downstream task (whether text generation, text translation, text classification or entity recognition.
\\In fact, the attention cell in the transformer aims at transforming the input embedding vector into three sub-vectors known as query, key, and value (See Fig. \ref{NNarch}(i)). This is performed by factoring the input vector \textit{x} by three parameterized matrices \textit{w\textsubscript{q}}, \textit{w\textsubscript{k}} and \textit{w\textsubscript{v}}, respectively. Then, a dot-product will be conducted between the query and the key vectors, followed by a Softmax scaling and another Softmax that will be then factored by the value vector and finally summed together.
\\Conditional Transformer Language \textit{CTRL} is an architecture based on the transformer that has been complemented with additional information such as style, content and task-specific features to control the generated text \cite{Keskar_etal_2019}.
\\The encoder-decoder transformer architecture has witnessed a dramatic expansion of variants. Some of these variants served visual contents such as Vision Encoder Decoder \textit{VED} \cite{OriginalVED} (integrated with CNN), Vision Language Pretraining \textit{VLP} \cite{OriginalVLP}, and Graph Transformer \textit{GTR} \cite{OriginalGTR}. Others served text generation purposes such as Longformer Encoder Decoder \textit{LED} \cite{OriginalLED}, Bidirectional and Auto-Regressive Transformer \textit{BART} \cite{Lewis_atal_2019} with its biomedical variant \textit{BioBART} \cite{Yuan_etal_2022}, Text-To-Text Transfer Transformer \textit{T5} \cite{OriginalT5} with its medical variant Label-To-Text Transformer \textit{LT3} \cite{Belkadi_atal_2023}, and Protein Language Model \textit{PLM} \cite{madani2023large}. VLP, in particular, has two well-known architectures called Contrastive Language Image Pretraining \textit{CLIP} \cite{LiCLIP_2021} and Bootstrapping Language Image Pretraining \textit{BLIP} \cite{li2022blip} with its medical model \textit{ClinicalBLIP} \cite{Ji2024ClinicalBLIP}. Both architectures have an image encoder along with a text decoder.

\subsubsection{Bidirectional Encoder Representation from Transformer (BERT)}
\textit{BERT} is an important architecture that has been built based on the transformer’s encoder \cite{Devlin_etal_2018}. BERT consists of two main components: \textit{pre-training} and \textit{fine-tuning}. The pre-training has taken only the encoder layer from the transformer architecture (See Fig. \ref{NNarch}(m)) and added two additional tasks, including masking and sentence prediction. Masking task aims at predicting the following word in a sentence, whereas sentence prediction aims to predict whether a sentence correlates to its following sentence. These tasks aim to make the model learn the general language characteristics. The second component of BERT is the fine-tuning, which is the part that will mimic the desired downstream task (\textit{i.e.}, text classification, entity detection, etc.). Note that BERT can be used for end-to-end classification where the entire model (pre-training and fine-tuning) will be updated and adjusted for a particular task, or it can be used as feature extraction where the output embedding of the pre-training component will be fed to another model like LSTM or RNN. In the feature extraction case, the model does need to be updated. In fact, BERT has already been pre-trained on large-scale text using BookCorpus (800 million words) and Wikipedia (2.5 billion words) sources. In addition, BERT has been pre-trained on a fixed vocabulary of 30,000 English words with a fixed embedding dimension of 768. This fixed vocabulary handles stemmed or root words where any derivational inflections such as \textit{ing} or \textit{ed} will be treated as individual tokens as \textit{\#\#ing} and \textit{\#\#ed} respectively (See Fig. \ref{NNarch}(l)). In this regard, the OOV problem has been significantly overcome.
\\Since the pretraining mechanism has led to computationally large models, which hinders the process of transferring such models, a concept of knowledge distillation has emerged to compress such large models into much smaller ones. Distillation consists of two models: teacher and student. The student model will be trained to mimic the larger model of the teacher. This can be performed by transferring the knowledge which usually lies in the hidden layers or so-called hint layers where the latent contextual information and relationships among the input features are depicted. This has led to the emergence of \textit{distilBERT} \cite{Sanh_etal_2019}.
\\On the other hand, BERT has been depicted in the medical domain by different pretrained models including \textit{PubMedBERT} \cite{OriginalPubMedBERT} and \textit{SMedBERT} \cite{OriginalSMedBERT}.

\subsubsection{Generative Pre-trained Transformer (GPT)}
\textit{GPT} is another architecture that has been built based on the transformer’s decoder, which is also known as \textit{Autoregressive}. GPT borrows the decoder component of the transformer architecture (See Fig. \ref{NNarch}(n)) to form a stack of decoders. Similar to BERT, GPT has a pretraining and fine-tuning part yet, the pre-training is mainly used for masked language modelling, where the following word in a sentence is being predicted with no sentence prediction. This pre-training is known as an unsupervised generative model or self-supervised, which refers to the task of predicting following words for the sake of automatic labelling. GPT has been pre-trained on BookCorpus (800 million words) and had 110 million parameters. Those statistics have been expanded with \textit{GPT-2}, \textit{GPT-3}, and \textit{GPT-4} (1.5 billion, 175 billion, and 1 trillion parameters, respectively). Similar to distilBERT, a smaller/compressed version of GPT has been depicted in \textit{distilGPT} \cite{Li_etal_2021_DistilGPT}.
\\GPT has been examined in terms of textual data augmentation where a Language Model-Based Data Augmentation \textit{LAMBADA} has been presented \cite{Anaby-Tavor_etal_2019}. LAMBADA was built based on a fine-tuned GPT-2 architecture in which the input is \textit{x} data of sentences along with their target \textit{y}. The aim of LAMBADA is to synthesize the sentences \textit{x} while preserving the target \textit{y}. Once the generation of new synthetic sentences is done, another layer of binary classification is conducted. This task is known as noise control or filtering, in which the GPT architecture will select the most accurate synthetic sentences. Such a selection will be based on the highest probability given by the classifier to a particular target \textit{y}. In other words, the sentences that have been classified with the highest probability in accordance with a specific class label will be selected.
\\Another GPT variant known as Differentially Private GPT \textit{DP-GPT} \cite{Al-Aziz_etal_2022,ZecevicETAL2024} has been depicted in the literature, which aims at providing a mathematical boundary by minimizing the logarithm of the probability of an individual. This can be performed by adding randomness or noise to avoid acquiring a higher probability of a target class (\textit{e.g.}, disease/diagnosis) for a particular individual (\textit{i.e.}, patient) and to ensure privacy.
\\Transformer Decoder with Memory-Compressed Attention \textit{T-DMCA} is a GPT that utilizes a decoder-only of the transformer architecture with a compressed memory that can handle longer sequences \cite{Liu_etal_2018_Generating}.
\\GPT  has been depicted in the medical domain by different pretrained models such as \textit{BioGPT} \cite{101093BioGPT}, \textit{Galactica} \cite{OriginalGalactica}, \textit{GPT-Neox} \cite{OriginalGPTneox}, \textit{CerebroGPT} \cite{oh2024use}. As well as, GPT has contributed toward the emergence of generative applications such as \textit{Meta Llama}, \textit{Google Bard}, \textit{Mistral}, \textit{Claude}, \textit{Gemma}, and \textit{ChatGPT}.

\subsubsection{Generative Adversarial Network (GAN)}
A recent architecture that has been examined for generating synthetic data is \textit{GAN}. This architecture consists of two main components: \textit{generator} and \textit{discriminator} (See Fig. \ref{NNarch}(j)). Generator will resemble a network that takes an input \textit{x} and attempts to predict/generate a synthetic copy of \textit{x} \cite{Goodfellow_etal_2014_GAN}. Whereas the discriminator is a network that takes input \textit{x} along with the synthetic \textit{x} generated by the generator network and tries to predict a binary class of correct or incorrect, corresponding to real or synthetic, respectively. In this regard, GAN will depict the \textit{cat and mouse} game in which the generator is producing new data and the discriminator is tested to identify which data is original and which data is synthetic. In this case, optimizing one network over the other will not be the ideal scenario. For example, an accurate generator means that the discriminator will be fooled with too many instances; vice versa, an accurate discriminator means that any generated data by the generator will be detected by the discriminator. Therefore, the ideal scenario is a trade-off between the accuracies of both networks (\(\approx 50\%\) for both networks). GAN is a concept where both feedforward (\textit{e.g.}, RNN) or convolutional (\textit{e.g.}, CNN) layers can be used either in the generator or in the discriminator.
\\GAN has been contributing toward alleviating the problem of exposure bias depicted in both RNN and LSTM, yet it is still facing some issues. First, there is a remarkable difficulty in terms of backpropagating from the discriminator to the generator. Second, since the generator is intended to create a realistic replica by inducing slight changes to the data, applying this to the discrete tokens (\textit{i.e.}, generating text) will cause a problem to the discriminator. This is because there will be no corresponding token for such slight changes in the limited dictionary space. Here comes the Sequence GAN or so-called \textit{SeqGAN}, which introduced the use of Reinforcement Learning \textit{RL} within GAN architecture \cite{Yu_etal_2016_SeqGAN}. RL is a new paradigm besides the supervised and unsupervised, where an agent is programmed to take an observational state from an environment (\textit{e.g.}, position within a game) and tries to predict an arbitrary value. Then, this arbitrary value predicted by the agent will be assessed to change the action within the environment \cite{Li_Yuxi_2017}. Let assume a chess game where the opponent makes a move; the agent will predict an arbitrary value that reflects a random movement. If this movement makes any benefit, the agent will be rewarded with a score (See Fig. \ref{NNarch}(g)). Otherwise, it will be punished with a penalty score. Hence, over time, the agent will learn to make accurate actions. RL has been widely used for mastering different games and has potential in the field of robotics. However, in the context of SeqGAN, the generator works as an agent, and the discriminator works as the evaluating environment. A remarkable problem known as mode collapse occurs due to the entire sequence evaluation conducted by the discriminator of GAN. Assessing an entire sequence would not be an accurate strategy, especially if the sequence is too long. Hence, an improved SeqGAN known as \textit{SegGAN} based on sub-sequence evaluation has been introduced \cite{Chen_etal_2019_SegGAN}.
\\Recently, due to the challenges facing GAN and VAE in terms of generating discrete variables such as text sequences, a combination of the two architectures has been introduced to form the Adversarially Regularized Autoencoder \textit{ARAE} \cite{Zhao_etal_2017_ARAE}. Such an architecture consists of the traditional layers of VAE, where input, encoder, coder, decoder, and output are depicted. Additionally, the resulting vector from the coding accompanied by sample distribution specified by the user will be fed as an input to a discriminator layer that will distinguish the two samples.

\pagebreak

\section{Appendix C: Evaluation Methods}
\label{sec:Evaluation}

\subsection{Similarity}
A good synthetic text should reflect some similarity against the original text in order to have a possible substitution between the two sets in the privacy-preserving scenarios and for data augmentation as well. In this section, a comprehensive discussion is provided on the distance-based similarity metrics that have been depicted in the literature.

\subsubsection{Bi-Lingual Evaluation Understudy (BLEU)}
The most common metric used by the literature in that sense is \textit{BLEU}. BLEU was originally proposed for evaluating machine translation text \cite{Papineni_etal_2002}. Recently, it gained high popularity due to the emergence of language generative models in which the generated text can be assessed in terms of its closeness to the original text. To understand BLEU, it is necessary to discuss the \textit{precision} which corresponds to the number of matching words within the synthetic text and the original text divided by the total number of words in the synthetic text which can be computed in the following equation:
\begin{equation}
\textrm{Precision} = \frac{\#Matching(W_s,W_o)}{W_s}\label{eq1}
\end{equation}
This can be computed for single matching terms or multiple consecutive terms which are known as \textit{N-grams}. N-grams refer to the number of consecutive words such as \textit{unigram} (\textit{i.e.}, single terms), \textit{bigram} (\textit{i.e.}, pairs of terms), \textit{trigram} (\textit{i.e.}, triples of terms), and \textit{quadgram} (\textit{i.e.}, quadruples of terms). Precision is also known as Positive Predictive Value \textit{PPV}. In fact, precision fails to estimate how accurate the synthetic text is since the repetition of matching words would lead to 100\% accuracy, meanwhile, the synthetic text would not make any sense. Therefore, a modified version of the precision known as \textit{Clipped precision} has emerged to consider additional factors of unique occurrences of terms along with the matching as depicted in the equation below:
\begin{equation}
P_n = \frac{\sum_{C \in (Candidates)}\sum_{ngram \in C}Count_{clip}(ngram)}{\sum_{\bar{C} \in (Candidates)}\sum_{\bar{ngram} \in \bar{C}}Count(\bar{ngram})}\label{eq2}
\end{equation}
Hence, for a sequence of text, each n-gram will be computed and multiplied to calculate the geometric average precision as in the equation below:
\begin{equation}
\textrm{Geometric Average Precision (N)} = exp \left( \sum\limits_{n=1}^{N} w_n \, log \, p_n \right)\label{eq3}
\end{equation}
Typically, \textit{N} is considered as 4 corresponding to the possible n-gram (\textit{i.e.}, unigram, bigram, trigram, quadgram), and \textit{w\textsubscript{n}} is the uniform weight which is equivalent to 1/N. One step before computing the BLEU is by addressing the length of the synthetic and original text in order to accurately estimate the similarity. This can be calculated as in the following equation:
\begin{equation}
\textrm{Brevity Penalty} =
\begin{cases}
    1,& \text{if } c > r\\
    e^{\frac{1-c}{r}},& \text{if } c\leq r
\end{cases}
\end{equation}
, where \textit{c} refers to the length of the synthetic text, and \textit{r} reflects the length of the original text. Now, BLEU can be calculated by multiplying the geometric average precision by the brevity penalty as in the following equation:
\begin{equation}
\textrm{BLEU} = BP*N\label{eq5}
\end{equation}
BLEU was used by multiple studies to compare the similarity between synthetic medical text and the original one \cite{Al-Aziz_etal_2022,amin-nejad_etal_2020,Begoli_etal_2018,Ive_etal_2020,Jing_etal_2018_Automatic,Kasthurirathne_etal_2019,Kasthurirathne_etal_2021,Li_etal_2021,Wang_etal_2019_artificial,WuETAL2022,Zhou_etal_2022_DataSifterText}. However, BLEU suffers from multiple drawbacks such as discarding the meaning of terms where synonyms could be treated incorrectly (e.g., medical $\neq$ clinical). This is also applied for any derivational inflection of the same term (e.g., clinical $\neq$ clinicals) \cite{Semeniuta_etal_2018}.
\\\textit{Self-BLEU} is another metric that extends the original BLEU to assess the diversity of synthetic text, where a one generated synthetic sentence can be assessed in terms of resembling the others \cite{Zhu_2018_Texygen}. Self-BLEU will calculate BLEU for every generated sentence, then compute the average of the entire generated synthetic document. The higher the value of Self-BLEU, the less diversity lies behind the synthetic text. In the context of evaluating synthetic medical text, this metric has been used by Guan et al. \cite{Guan_etal_2018,Guan_etal_2019}.
\\While BLEU is considering only the precision, there is another metric that can assess the retrieval terms, which is known as \textit{Recall} or \textit{Sensitivity} and sometimes as Negative Predictive Value \textit{NPV}. Recall corresponds to the number of matching words between the original \textit{W\textsubscript{o}} and synthetic \textit{W\textsubscript{s}} text divided by the total number of words within the original text as in the following equation:
\begin{equation}
\textrm{Recall} = \frac{\#Matching(W_s,W_o)}{W_o}\label{eq6}
\end{equation}
Therefore, a new version of BLEU introduced by Google, which is known as \textit{GLEU} \cite{Wu_etal_2016_GLEU}. GLEU is simply taking the minimum of calculated precision and recall between the generated and original text. GLEU has the same performance as BLEU in corpus-level and better performance in the sentence-level. GLEU has been used by the study of Kasthurirathne et al. \cite{Kasthurirathne_etal_2019}.

\subsubsection{Recall Oriented Understudy for Gisting Evaluation (ROUGE)}
\textit{ROUGE} is another metric used to evaluate synthetic text. ROUGE has been widely used for evaluating machine translation and automatic text summarization \cite{Lin_Chin-Yew_2004_ROUGE}. Similar to BLEU, ROUGE employs the precision for multiple n-grams where \textit{ROUGE-1} will focus on unigram, \textit{ROUGE-2} for bigram, and so on. Additionally, ROUGE can extend the evaluation to include recall and \textit{F1-score}, which is a harmony between precision and recall and computed as in the following equation:
\begin{equation}
\textrm{F1-score} = 2* \frac{Precision * Recall}{Precision + Recall}\label{eq7}
\end{equation}
In the context of evaluating synthetic medical text, ROUGE has been used by multiple studies \cite{amin-nejad_etal_2020,Jing_etal_2018_Automatic,Libbi_etal_2021,Liu_etal_2018,WuETAL2022}. Another modified version of ROUGE known as \textit{ROUGE-L} has emerged to extend the matching analysis through the Longest Common Subsequence \textit{LCS}. LCS refers to the longest consecutive matching words between the original and synthetic text even if such consecutive sequence has been disrupted by different words. ROUGE-L has been used for synthetic medical text evaluation by Amin-Nejad et al. \cite{amin-nejad_etal_2020} and Ive et al. \cite{Ive_etal_2020}.

\subsubsection{Metric for Evaluation of Translation with Explicit Ordering (METEOR)}
\textit{METEOR} is another metric used for evaluating synthetic text. Similar to both BLEU and ROUGE, METEOR has been extensively addressed for evaluating machine translation and automatic text summarization or generation. It employs precision and recall between the original and synthetic text with only unigram matching. However, the harmonic mean of METEOR is slightly different in which the recall is highly weighted over precision, as in the following equation \cite{Denkowski_Lavie_2014}:
\begin{equation}
F_{mean} = 10* \frac{Precision * Recall}{Recall + (9*Precision)}\label{eq8}
\end{equation}
In the context of synthetic medical text evaluation, METEOR has been used by multiple studies \cite{Begoli_etal_2018,Jing_etal_2018_Automatic}.

\subsubsection{Translation Edit Rate (TER)}
\textit{TER} is a simple metric that examines the least number of operations required to transform a synthetic text into the original one. It can be applied over character-level (identifying the number of character edits to transform a word into another) \cite{Wang_etal_2016_TER} or over word-level (identifying the number of word edits to transform a sentence into another) \cite{Snover_etal_2006}. Assuming the word-level, TER measures the ratio between the number of word edits W\textsubscript{edits} over the average number of words within the original text W\textsubscript{o} as depicted in the following equation:
\begin{equation}
\textrm{TER} = \frac{W_{edits}}{Avg(W_o)}\label{eqTER}
\end{equation}
TER metric has been depicted by the study of Ive et al. \cite{Ive_etal_2020} where synthetic mental health text was evaluated against the original text. In addition, Wu et al. \cite{WuETAL2022} used TER to evaluate generated synthetic text for summarizing medical reports’ tables compared to human-generated summarizations.

\subsubsection{Cosine and Jaccard}
On the other hand, there are metrics that examine the similarity among word vectors. In this type of evaluation, the similarity between words is performed by converting the words themselves into vectors. These vectors are either formed through the character occurrence or via word embedding. The first and most popular vector-based metric is the \textit{Cosine similarity}, where the cosine of the angle between two vectors would resemble the similarity among them. Let assume two vectors \textit{A} and \textit{B}, the cosine is calculated as in the following equation:
\begin{equation}
Cosine(A,B) = \frac{A*B}{\left\|A\right\|\left\|B\right\|}\label{eq10}
\end{equation}
Melamud \& Shivade \cite{Melamud_shivade_2019} have used the cosine to compute the similarity between a synthetic medical and real text. Similarly, \textit{Jaccard} is another vector-based metric that examines the similarity between two vectors \textit{A} and \textit{B} through the ratio of intersection between them divided by the union of the two vectors as in the following equation:
\begin{equation}
Jaccard(A,B) = \frac{\lvert A \cap B \rvert}{\lvert A \cup B \rvert}\label{eq11}
\end{equation}
Al Aziz et al. \cite{Al-Aziz_etal_2022} have used Jaccard to estimate the similarity between a synthetic medical text and original text.

\subsubsection{Consensus-based Image Description Evaluation (CIDEr)}
\textit{CIDEr} is another metric that examines the similarity between two sentences \textit{C} and \textit{R} through applying n-gram cosine similarity among the individual term-vector within the two sentences \cite{Vedantam_etal_2014}. To form the word vector, CIDEr utilizes the Term Frequency – Inverse Document Frequency \textit{TF-IDF} where a term \textit{t} within the generated text will be addressed in terms of frequency over the original text multiplied by logarithm of the number of sentences \textit{N} divided by the number of sentences \textit{N\textsubscript{t}} that contain \textit{t} as in the following equation \cite{Robertson_Stephen_2004}:
\begin{equation}
TF-IDF(t) = TF(t)*log\frac{N}{N_t}\label{eq12}
\end{equation}
Hence, CIDEr will consider the average cosine similarity as in the following equation:
\begin{equation}
CIDEr(C,R) = \sum\frac{g^n(C)*g^n(R)}{\left\|g^n(C)\right\|\left\|g^n(R)\right\|}\label{eq13}
\end{equation}
\textit{g\textsuperscript{n}(C)} is the TF-IDF vector formed by the n-gram words from the first sentence, and \textit{g\textsuperscript{n}(R)} is the TF-IDF vector formed by the n-gram words from the second sentence. Jing et al. \cite{Jing_etal_2018_Automatic} and Lee \cite{Lee_2018_Natural} have used CIDEr to estimate the similarity between a synthetic medical text against the original one.
\\Similar to TFIDF, Best Match \textit{BM25} adds tunable parameters with a smoothed log to get probabilistic and non-linear similarity ranking \cite{robertson2009probabilistic}.
\\Some studies have addressed the role of \textit{word-level feature selection} among the original and synthetic medical text \cite{Kasthurirathne_etal_2019,Kasthurirathne_etal_2021}. The overlapping between selected features/words has then been examined.

\subsubsection{Semantic-based Metrics}
All the aforementioned metrics are based on word-level similarity; there are metrics that consider the semantic similarity, such as the \textit{BERTscore} \cite{zhang2019bertscore}, \textit{SemScore} \cite{aynetdinov2024semscore}, and \textit{F1cXb} \cite{parres2024improving}, which are metrics that use the BERT architecture to compute how similar two sentences are based on the sum of cosine similarity of their BERT token embeddings. Another metric is Semantic Propositional Image Caption Evaluation \textit{SPICE} \cite{anderson2016spice} which also compute the semantic similarity between two sentences based on tuples of objects (\textit{e.g., nouns}), attributes (\textit{i.e.}, adjectives) and relations (\textit{i.e.}, verbs and prepositions).
\\Another medically-specific metric is the Medical Conceptual Similarity (MEDCON) which aims to compute the similarity between two sentences based on identifying and comparing medical concepts \cite{YimETAL2023}.

\subsubsection{Human Assessment}
Some studies have addressed the Human Assessment for similarity evaluation where multiple annotators are asked to label synthetic medical text. Then, the Inter-Annotator Agreement \textit{IAA} is computed through \textit{Cohen’s Kappa} which can be calculated as in the following equation \cite{McHugh_Mary_2012}:
\begin{equation}
\textrm{Kappa} = \frac{P_o-P_e}{1-P_e}\label{eq14}
\end{equation}
, where \textit{P\textsubscript{o}} is the relative observed agreement among the annotators, and \textit{P\textsubscript{e}} is the hypothetical probability of chance agreement. The study of Brekke et al. \cite{Brekke_etal_2021} and Rama et al. \cite{Rama_etal_2018} have used human assessment to evaluate the similarity of synthetic clinical text against the original.
\\In the same context, Evaluator Reliability Error \textit{ERE} is another metric to compute the error resulting from different annotators \cite{Guan_etal_2018}.

\subsection{Privacy}
Apart from the similarity, medical-generated synthetic text should be secured from threats like re-identification or membership inference. Having secured synthetic medical text will facilitate having publicly available medical free-text and enable sharing among institutions, which will open the door for significant contributions. The simplest approach to evaluate security was depicted by the study of Lee \cite{Lee_2018_Natural}, where the authors have checked for the existence of PII or PHI within the generated synthetic text. Yet, the non-existence of such sensitive identifiers will not obstruct the threats where some attacks used implicit features to re-identify individuals.
\\Another approach depicted by the literature is to examine the similarity between synthetic individual records against the original records. Zhou et al. \cite{Zhou_etal_2022_DataSifterText} have used BLEU (See Equation \ref{eq5}) and cosine (See Equation \ref{eq10}) in a reverse manner in which the high similarity indicates less privacy. Other studies like Kasthurirathne et al. \cite{Kasthurirathne_etal_2019,Kasthurirathne_etal_2021} have used the \textit{Hamming distance}, which is a metric similar to TER (See Equation \ref{eqTER}) in terms of calculating the number of required operations to transform a sentence into another, for the same purpose.
\\Much more sophisticated metrics for privacy assessment have been depicted in the literature. These metrics provide a mathematical boundary to check whether implicit information of an individual still exists within the synthetic text. The following subsections will depict such metrics.

\subsubsection{Standard Log Likelihood Ratio Test ($G^2$)}
This statistical test can be employed to address the membership of certain words toward a set of text \cite{Rayson_etal_2004}. In this regard, it can be used to calculate the statistical significance of a term in accordance with either the original or synthetic text. The higher probability of a term attained in respect to a text set means that this term has been taken from such a set. Therefore, smaller value of \textit{G\textsuperscript{2}} indicates uncertainty of membership. Let \textit{E1} and \textit{E2}  be the two corpora of original and synthetic text, respectively. The expected frequency of a term \textit{t} for both corpora can be calculated as in the following equations:
\begin{equation}
\textrm{E1} = c* \frac{a+b}{c+d}\label{eq15}
\end{equation}
\begin{equation}
\textrm{E2} = d* \frac{a+b}{c+d}\label{eq16}
\end{equation}
where \textit{a} and \textit{b} correspond to the number of occurrences for \textit{t} in the original and synthetic corpora respectively. Whereas \textit{c} and \textit{d} correspond to the total number of words within the original and synthetic corpora, respectively. Hence, \textit{G\textsuperscript{2}}  can be computed as in the following equation:
\begin{equation}
G^2 = 2* \left( a\,ln\, \left( \frac{a}{E1} \right) + b\, ln\, \left( \frac{b}{E2} \right) \right)\label{eq17}
\end{equation}
Considering the difference in corpora sizes, it is necessary to examine the effect size through the following equation:
\begin{equation}
\textrm{Effect Size} = \frac{G^2}{c+d} * ln\, min(E1,E2)\label{eq18}
\end{equation}
\textit{G\textsuperscript{2}} has been used by the study of Al Aziz et al. \cite{Al-Aziz_etal_2022} to assess the privacy of synthetic medical text.

\subsubsection{Negative Log Likelihood (NLL)}
\textit{NLL} is another metric examined for privacy. This evaluation test has been proposed by Yu et al. \cite{Yu_etal_2016_SeqGAN} to assess the proposed SeqGAN method for text generation. It aims to estimate how bad the generative model is by examining the loss through deriving the negative natural logarithm of the likelihood. The better the prediction is, the lower value of NLL is obtained. Hence, it can be used to assess the generative model towards predicting sequences of words from the real data as in the following equation:
\begin{equation}
NLL_{test} = - E_{Y_{1:T\sim G_{real}}} \left[ \sum\limits_{t=1}^{T} log \left( G_\theta (y_t\mid Y_{1:t-1}) \right) \right]\label{eq19}
\end{equation}
where \textit{G\textsubscript{real}} is the distribution of real data word, and \textit{G\textsubscript{$\theta$}} is the generative model. NLL has been used by the study of Guan et al. \cite{Guan_etal_2018,Guan_etal_2019}.

\subsubsection{Perplexity (PPL)}
\textit{PPL} is another metric that aims to intrinsically evaluate a language model for better parameter tuning. Given a real dataset containing sequences of words, a good language generative model, if tested on these words, would give high probability. This means that such words are familiar to the model. However, as the test data increases, the probability will start to drop. Hence, the perplexity emerged as a size-independent evaluation metric by normalizing the probability over the total number of words within the test set as in the following equation \cite{Keselj_Vlado_2009}:
\begin{equation}
PPL(W) = \sqrt[\leftroot{-2}\uproot{2}N]{\frac{1}{P(W_1,W_2,\dotsc,W_N)}}\label{eq20}
\end{equation}
where \textit{N} is the number of words within the test set. The lower the perplexity value, refers to a good model. PPL has been used by multiple studies to evaluate the privacy of synthetic medical text \cite{amin-nejad_etal_2020,Ive_etal_2020,Melamud_shivade_2019}.

\subsubsection{Differentially Training Privacy (DTP)}
While both PPL and NLL are restricted to deep learning architectures, \textit{DTP} is a metric that employs a local property of any classification model to examine how such a model would reveal sensitive information \cite{Long_etal_2017}. Given a model \textit{M} which has possible prediction targets \textit{Y} and a training set \textit{T}, let \textit{t} be an individual record that belongs to \textit{T}. In order to preserve the privacy of \textit{t}, it is necessary to examine the prediction probability of \textit{t} using a model that trained with \textit{t} and a model that trained without \textit{t}. As long as the difference is minimized, \textit{t} will be protected as depicted in the following equation:
\begin{equation}
PDTP_{M,T}(t) = max_{y\in Y} \left( \mid log\, p_{M(T)}(y\mid t) - log\, p_{M(T\setminus \{t\})}(y\mid t) \mid \right)\label{eq21}
\end{equation}
where \textit{M(T)} is the model that has been trained with \textit{t}, and \textit{$M(T\setminus \{t\})$} is the model that trained without \textit{t}. Whereas, \textit{P($y\mid t$)} is the probability of predicting the target class of \textit{t} record. Melamud \& Shivade \cite{Melamud_shivade_2019} have presented a Sequential Point-wise Differentially Training Privacy \textit{S-PDTP} to suit the case of language model generation. In this regard, the individual \textit{t} has been considered as the sequence of words $w^{\mathrm{c}}_{i}\mid w^{\mathrm{c}}_{1\dotsc i-1}$ appeared in an individual clinical note, which can be computed as in the following equation:
\begin{multline}
S-PDTP_{M,T}(c) = max_{i\in 1\dotsc \mid c\mid} \left( \mid log\, p_{M(T)}w^{\mathrm{c}}_{i}\mid w^{\mathrm{c}}_{1\dotsc i-1} - log\, p_{M(T\setminus \{c\})}w^{\mathrm{c}}_{i}\mid w^{\mathrm{c}}_{1\dotsc i-1} \mid \right)\label{eq22}
\end{multline}
\\Another metric known as \textit{Privacy Budget} is depicted through the literature, which aims to calculate epsilon \textepsilon  that is intended to indicate the privacy loss allowed for querying a dataset. More noise added to the data instances would lead to lower privacy loss which indicates better privacy \cite{ZecevicETAL2024}.
\\Some studies have addressed the Human Assessment for privacy evaluation where multiple annotators are asked to label synthetic medical text into reveal-identify and no-reveal, then, computing IAA with Cohen’s Kappa. The study of Libbi et al. \cite{Libbi_etal_2021} has used this type of evaluation to assess the privacy of the synthetic text; multiple interesting findings have been found. First, generated synthetic text might produce real identities, yet it presents them with incoherence and fake details (\textit{e.g.}, combining different names). This could provide a natural privacy-protecting feature. However, the synthetic text copied long sequences associated with real medications in some cases. If such medications are unique to a particular individual, it might harm the privacy. Therefore, exact copying of longer sequences within the synthetic text must be reviewed carefully.

\subsection{Structure}
Some studies have paid much attention to evaluating the coherence of the synthetic medical text. Lee \cite{Lee_2018_Natural} has used the \textit{Odd ratio}, which is a statistical measure to examine the word distribution among the original and synthetic text. It is calculated in the following equation:
\begin{equation}
\textrm{Odd ratio(w)} = \frac{D_{wi}\setminus R}{D_{wj}\setminus S}\label{eq23}
\end{equation}
, where \textit{D\textsubscript{wi}} and \textit{D\textsubscript{wj}} are the number of occurrences of word \textit{w}, while \textit{R} and \textit{S} refer to the total number of words within the original and synthetic corpora, respectively. In this essence, the authors selected some terms such as \textit{pregnancy} or \textit{overdose} and examined them with other criteria such as age or gender in terms of their distribution across the original and the synthetic text. Results showed a satisfactory distributional resemblance between the original and synthetic sets. For example, the root word of \textit{pregnancy} as \textit{preg} has occurred with females 134 times within the original text and zero occurrences with males. This has been depicted within the synthetic text in which the occurrence of such a root-word was 44 times with females and zero with males.
\\\textit{Clustering} is another method that has been used to assess the semantic diversity \cite{Zhou_etal_2022_DataSifterText,Modersohn_etal_2022}.
\\On the other hand, \textit{pattern match} was used specifically to assess the generation of text towards annotation where simple word count can be used to match annotated tags. Such a word count metric has also helped the assessment of sentences length \cite{HiebelETAL2023,FreiETAL2022}.

\subsubsection{Adversarial Success (AdvSuc)}
\textit{AdvSuc} is a much sophisticated test which is very similar to \textit{Turing test} in which the synthetic and original text are tested in terms of real and synthetic. However, this test has been employed by an individual classifier usually known as a discriminator within the GAN architecture. Once the synthetic is being generated by the generator part, both the synthetic and original text will be passed into the discriminator to be tested. The discriminator will be trained to distinguish between real and synthetic text. The ideal scenario is when the discriminator obtained 50\% of accuracy, meaning that it can be confused between real and synthetic, which reflects a high quality of synthetic text. All the studies that have utilized GAN architecture of synthetic medical text generation were considering AdvSuc within their experiments \cite{Al-Aziz_etal_2022,Guan_etal_2018,Guan_etal_2019,Zhou_etal_2022_DataSifterText}.

\subsubsection{Natural Language Inference (NLI)}
For contradiction, Negation correctness has been depicted in the literature where a simple search for negation terms is conducted and reviewed in terms of the correctness. \textit{NLI} is an automatic task used to evaluate the contradiction of synthetic text. This task takes two sentences and attempts to give a label whether the sentences are contradicting, entailing or neutral \cite{Melamud_shivade_2019}.

\subsubsection{Transfer Learning-Based BLEU (BLEURT)}
In order to assess fluency, \textit{BLEURT} \cite{BLEURT} has been depicted in the literature. Such a metric is based on the BERT architecture, which aims to accommodate a regression task (\textit{i.e.}, predicting a score) by taking two sentences (\textit{i.e.}, synthetic and original) as input, to indicate whether a generated sentence is fluent and conveys the meaning of the original one \cite{YimETAL2023,SchlegelETAL2023}.
\\It is worth mentioning that PPL is used also to assess the structure of the synthetic text where some studies used it to evaluate the accuracy of the generative model itself \cite{Spinks_etal_2018,Liu_etal_2018}.
\\Lastly, the Human Assessment with IAA and ERE have been used as well to assess the structure of the generated synthetic text.

\subsection{Utility}
In this evaluation, the usefulness of the generated synthetic medical text is examined. This can be depicted through a downstream NLP task. In the following, these tasks will be explained.

\subsubsection{Text Classification}
This is one of the popular tasks that aims at classifying a set of text into a predefined class label. In the medical context, \cite{NingsihETAL2022} have addressed this downstream task as a utility for classifying COVID news.

\subsubsection{Disease Classification}
This task aims to classify the patients records and healthy controls into a binary class depending on a particular disease \cite{Kasthurirathne_etal_2019,LitakeETAL2024,Abdollahi_etal_2020,Al-Aziz_etal_2022}.

\subsubsection{Phenotype Classification}
This task aims to classify patient records into one or more predefined symptoms \cite{Velichkov_etal_2020,Igarashi_Nihei_2022,LorgeETAL2024,amin-nejad_etal_2020,Ive_etal_2020}.

\subsubsection{Diagnosis Prediction}
This task aims to process patient records to predict the level/progression of a specific disease (\textit{e.g.}, mild, moderate, or severe) \cite{ScrogginsETAL2024,LatifETAL2024,Lee_2018_Natural,BarretoETAL2023}.

\subsubsection{Readmission Prediction}
This task aims to process patient records and attempt to predict whether this patient will be readmitted again or not \cite{amin-nejad_etal_2020,Lu_etal_2021_Textual,AminNejadETAL2020}.

\subsubsection{Named Entity Recognition (NER)}
This task aims to extract medical entities such as medications, diseases, lab test or symptoms \cite{FreiETAL2022,Li_etal_2021,HiebelETAL2023,TangETAL2023,AggarwalETAL2023}.

\subsubsection{Adverse Drug Reaction/Event Extraction (ADR/ADE)}
This is a sub-task of NER which narrows the extraction to include mentions of adverse event or reaction through the medical text \cite{Tao_etal_2019,Ishikawa_etal_2022}

\subsubsection{Relation Extraction (RL)}
This is a complementary task to NER where the aim is to identify the relations among medical entities extracted by NER such as protein-protein interaction, drug-drug interaction and chemicals induced diseases \cite{Brekke_etal_2021,PengETAL2023,DelmasETAL2024,XuETAL2024}.

\subsubsection{Automatic De-identification}
This task is also known as \textit{Safe Harbor} where the aim is to replace PIIs/PHIs with random codes. De-identification is sometimes conducted by adding noise to PII/PHI in such a way that disable the re-identification. However, the manual de-identification would seem cumbersome and time-consuming task. Therefore, multiple studies have attempted to automate the de-identification as a NER task where a model is trained to identify PII/PHI and replace them with secured surrogates (\textit{e.g.}, replacing John by Name) \cite{Libbi_etal_2021,SinghETAL2024,LundETAL2024}.

\subsubsection{Question Answering (QA)}
This task aims to extract an exact portion of text that satisfies a typed question. This task becomes too valuable in the medical domain where it serves medical consultations \cite{QUETAL2023,ReynoldsETAL2024,KotschenreutherETAL2024,NguyenETAL2023}.

\subsubsection{Report Generation and Text Summarization}
This task aims at generating text for a particular aim. Some studies have used this task to process medical images (\textit{e.g.}, X-Ray) and attempt to generate synthetic text to describe the diagnosis \cite{Spinks_etal_2018,Jing_etal_2018_Automatic}. Whereas the study of Liu \cite{Liu_etal_2018} have used this task to autocomplete clinical notes. Lastly, Wu et al. \cite{WuETAL2022} have used this task by generating medical report’s table summarization.

\pagebreak

\begin{landscape}
    \section{Appendix D: Selected Articles}
\label{sec:SelectedArticles}
\tiny
\begin{longtblr}[caption = {Summary of selected articles}, label= {LR}]{colspec = {p{0.5cm}p{0.5cm}p{2cm}p{2cm}p{3cm}p{2cm}p{1cm}p{3cm}p{2cm}p{2cm}p{2cm}}, rowhead=1}
\hline
Article                              & Year & Purpose                               & Generation Method                 & Approach / Architecture                                          & Data Source                                                 & Language            & Evaluation Method                                                                                      & Evaluation Paradigm                             & Utility                                                           & Clinical Source                                         \\
\hline
\cite{GoldsteinAyelet2015}           & 2015 & Assistive Writing                     & Knowledge Source, Text Processing & Template-based                                                   & Publicly Available                                            & English           & Human Assessment                                                                                       & Structure                                       & Text Summarisation                                                & Discharge Summaries                                     \\
\cite{Spinks_etal_2018}              & 2018 & Assistive Writing                     & Neural Network                    & ARAE                                                             & Publicly Available (IU X-RAY)                                 & English           & PPL                                                                                                    & Structure, Test on Utility                      & Report Generation                                                 & Radiology                                               \\
\cite{Guan_etal_2018}                & 2018 & Privacy-preserving                    & Neural Network                    & SeqGAN (LSTM as generator, CNN and Bi-LSTM as discriminator), RL & Private EHR/EMR                                               & Chinese           & BLEU, NLL, AdvSuc, Human Assessment (ERE)                                                              & Similarity, Structure, Test on Utility          & Disease Classification                                            & Discharge Summaries                                     \\
\cite{Jing_etal_2018_Automatic}      & 2018 & Assistive Writing                     & Neural Network                    & Seq2Seq (CNN Encoder and LSTM Decoder)                           & Publicly Available (IU X-RAY)                                 & English           & BLEU, ROUGE, CIDEr, METEOR                                                                             & Similarity, Test on Utility                     & Report Generation                                                 & Radiology                                               \\
\cite{Liu_etal_2018}                 & 2018 & Assistive Writing                     & Neural Network                    & Transformers (TDMCA)                                             & Publicly Available (MIMIC-III)                                & English           & PPL, ROUGE, Human Assessment                                                                           & Similarity, Structure, Test on Utility          & Report Generation                                                 & Discharge Summaries                                     \\
\cite{Begoli_etal_2018}              & 2018 & Corpus Building                       & Text Processing, Knowledge Source & UML, Template-based                                              & Publicly Available (MIMIC-III)                                & English           & BLEU, METEOR                                                                                           & Similarity                                      & -                                                                 & Discharge Summaries                                     \\
\cite{Lohr_etal_2018}                & 2018 & Corpus Building                       & Text Processing                   & Template-based                                                   & Online source                                                 & German            & Human Assessment                                                                                       & Structure                                       & -                                                                 & Clinical Practice Guidelines                            \\
\cite{Lee_2018_Natural}              & 2018 & Privacy-preserving, Augmentation      & Neural Network                    & Seq2Seq (LSTM Encoder and LSTM Decoder)                          & Private EHR/EMR                                               & English           & BLEU, ROUGE, Odd-ratio, CIDEr, Human Assessment (Check Sensitive Information), Classification Accuracy & Similarity, Structure, Privacy, Test on Utility & Diagnose Prediction                                               & Discharge Summaries                                     \\
\cite{Rama_etal_2018}                & 2018 & Corpus Building, Annotation           & Manual                            & -                                                                & Private EHR/EMR                                               & Norwagian         & Human Assessment (IAA), Pattern Match, Classification Accuracy                                         & Structure, Test on Utility                      & Named Entity Recognition, Relation Extraction                     & History of Present Illness (HPI)                        \\
\cite{Wang_etal_2019_artificial}     & 2019 & Usefulness                            & Neural Network                    & Transformers (CTRL), RAKE                                        & Publicly Available (MIMIC-III)                                & English           & BLEU, ROUGE                                                                                            & Similarity, Test on Utility                     & Phenotype Classification, Relation Extraction                     & Discharge Summaries                                     \\
\cite{Melamud_shivade_2019}          & 2019 & Privacy-preserving                    & Neural Network                    & LSTM                                                             & Publicly Available (MIMIC-III)                                & English           & PPL, PDTP, Human Assessment, NLI                                                                       & Similarity, Structure, Privacy                  & -                                                                 & Discharge Summaries                                     \\
\cite{Kasthurirathne_etal_2019}      & 2019 & Usefulness, Privacy-preserving        & Neural Network                    & SeqGAN                                                           & Private EHR/EMR                                               & English           & BLEU, Hamming Distance                                                                                 & Similarity, Privacy, Test on Utility            & Disease Classification                                            & Patient Laboratory Test                                 \\
\cite{Tao_etal_2019}                 & 2019 & Augmentation                          & Knowledge Source                  & Gazetteers                                                       & Publicly Available                                            & English           & Classification Accuracy                                                                                & Test on Utility                                 & Adverse Drug Reaction/Event Extraction                            & Drug and Medication                                     \\
\cite{KurisinkelETAL2019}            & 2019 & Assistive Writing                     & Neural Network                    & Seq2Seq (RNN Encoder and RNN Decoder)                            & Publicly Available (MIMIC-III)                                & English           & BLEU, Human Assessment                                                                                 & Similarly, Structure                            & Report Generation                                                 & Discharge Summaries                                     \\
\cite{PengETAL2019}                  & 2019 & Assistive Writing                     & Neural Network                    & GPT-2                                                            & Private EHR/EMR                                               & Chinese           & Human Assessment                                                                                       & Similarity                                      & Report Generation                                                 & Discharge Summaries                                     \\
\cite{LiETAL2019}                    & 2019 & Assistive Writing                     & Neural Network                    & GTR, CNN (DenseNet)                                              & Publicly Available (IU X-RAY)                                 & English           & BLEU, ROUGE, CIDEr, Human Assessment, Classification Accuracy                                          & Similarity, Structure, Test on Utility          & Disease Classification                                            & Radiology                                               \\
\cite{AminNejadETAL2020}             & 2020 & Usefulness                            & Neural Network                    & Transformers (Encoder-Decoder), GPT-2 (Decoder)                  & Publicly Available (MIMIC-III)                                & English           & Classification Accuracy                                                                                & Test on Utility                                 & Phenotype Classification, Readmission Prediction                  & Discharge Summaries                                     \\
\cite{BorchertETAL2020}              & 2020 & Corpus Building                       & Knowledge Source, Text Processing & UML, SpaCy                                                       & Online source                                                 & German            & Human Assessment (IAA), Pattern Match                                                                  & Structure                                       & -                                                                 & Clinical Practice Guidelines                            \\
\cite{NishinoETAL2020}               & 2020 & Assistive Writing, Augmentation       & Neural Network                    & Seq2Seq (GRU Encoder and GRU Decoder), RL, BERT                  & Publicly Available (MIMIC-CXR)                                & Japanese, English & BLEU, ROUGE, Human Assessment                                                                          & Similarity, Structure, Test on Utility          & Report Generation                                                 & Radiology                                               \\
\cite{Velichkov_etal_2020}           & 2020 & Corpus Building                       & Manual                            & -                                                                & Private EHR/EMR                                               & Bulgarian         & Classification Accuracy                                                                                & Test on Utility                                 & Phenotype Classification                                          & Discharge Summaries                                     \\
\cite{Abdollahi_etal_2020}           & 2020 & Augmentation                          & Knowledge Source, Neural Network  & WordNet, GloVe, word2vec, BioWord2Vec                            & Publicly Available                                            & English           & Classification Accuracy                                                                                & Test on Utility                                 & Disease Classification                                            & Discharge Summaries                                     \\
\cite{amin-nejad_etal_2020}          & 2020 & Augmentation                          & Neural Network                    & Transformers, GPT-2                                              & Publicly Available (MIMIC-III)                                & English           & PPL, BLEU, ROUGE                                                                                       & Similarity, Test on Utility                     & Readmission Prediction, Phenotype Classification                  & Discharge Summaries                                     \\
\cite{Ive_etal_2020}                 & 2020 & Privacy-preserving, Usefulness        & Neural Network                    & Transformers (CTRL)                                              & Private EHR/EMR                                               & English           & BLEU, PPL, TER, Human Assessment (IAA), Classification Accuracy                                        & Similarity, Structure, Privacy, Test on Utility & Phenotype Classification                                          & Discharge Summaries                                     \\
\cite{Kang_etal_2020}                & 2021 & Augmentation                          & Text Processing, Knowledge Source & UML, EDA                                                         & Publicly Available                                            & English           & Classification Accuracy                                                                                & Test on Utility                                 & Named Entity Recognition                                          & Population, Intervention, Comparison, and Outcome (PICO)\\
\cite{ZhouETAL2021}                  & 2021 & Assistive Writing                     & Neural Network                    & Seq2Seq (DenseNet CNN Encoder and LSTM Decoder)                  & Publicly Available (IU X-RAY, MIMIC-CXR)                      & English           & BLEU, ROUGE, CIDEr, METEOR, Hamming Distance                                                           & Similarity, Test on Utility                     & Report Generation                                                 & Radiology                                               \\
\cite{KagawaETAL2021}                & 2021 & Privacy-preserving, Corpus Building   & Manual                            & crowdsourcing human-in-the-loop                                  & Private EHR/EMR                                               & Japanese          & Human Assessment                                                                                       & Similarity, Structure, Privacy                  & -                                                                 & Patient Laboratory Test                                 \\
\cite{ChintaguntaETAL2021}           & 2021 & Augmentation, Usefulness              & Neural Network                    & GPT-3                                                            & Manual Collection and Curation                                & English           & ROUGE, Negation Correctness, Human Assessment                                                          & Similarity, Structure, Test on Utility          & Text Summarisation                                                & Medical Conversations                                   \\
\cite{Guan_etal_2019}                & 2021 & Privacy-preserving                    & Neural Network                    & SeqGAN (LSTM as generator, CNN and Bi-LSTM as discriminator), RL & Private EHR/EMR                                               & Chinese           & BLEU, AdvSuc, NLL, Human Assessment (ERE)                                                              & Similarity, Structure, Test on Utility          & Disease Classification                                            & Discharge Summaries                                     \\
\cite{Li_etal_2021}                  & 2021 & Usefulness                            & Neural Network                    & Transformers (CTRL), GPT-2, CharRNN, SeqGAN                      & Publicly Available                                            & English           & BLEU                                                                                                   & Similarity, Structure, Test on Utility          & Named Entity Recognition                                          & History of Present Illness (HPI)                        \\
\cite{Abdollahi_etal_2021}           & 2021 & Augmentation                          & Knowledge Source                  & GloVe, WordNet, UMLs                                             & Publicly Available                                            & English           & Classification Accuracy                                                                                & Test on Utility                                 & Disease Classification                                            & Discharge Summaries                                     \\
\cite{Brekke_etal_2021}              & 2021 & Augmentation                          & Manual                            & -                                                                & Private EHR/EMR                                               & Norwegian         & Classification Accuracy                                                                                & Test on Utility                                 & Named Entity Recognition, Relation Extraction                     & History of Present Illness (HPI)                        \\
\cite{Libbi_etal_2021}               & 2021 & Annotation, Privacy                   & Neural Network                    & LSTM, GPT-2                                                      & Private EHR/EMR                                               & Dutch             & ROUGE, BM25, Human Assessment (IAA), Pattern Match                                                     & Similarity, Structure, Test on Utility          & De-Identification                                                 & Discharge Summaries                                     \\
\cite{Kasthurirathne_etal_2021}      & 2021 & Usefulness, Privacy-preserving        & Neural Network                    & SeqGAN                                                           & Private EHR/EMR                                               & English           & BLEU, GLEU, Hamming Distance                                                                           & Similarity, Privacy, Test on Utility            & Disease Classification                                            & Patient Laboratory Test                                 \\
\cite{Lu_etal_2021_Textual}          & 2021 & Augmentation                          & Neural Network                    & distil-GPT-2, LAMBADA                                            & Publicly Available (MIMIC-III)                                & English           & Classification Accuracy                                                                                & Test on Utility                                 & Readmission Prediction                                            & Discharge Summaries                                     \\
\cite{Issifu_Ganiz_2021}             & 2021 & Augmentation                          & Text Processing, Knowledge Source & UML, EDA                                                         & Publicly Available                                            & English           & Classification Accuracy                                                                                & Test on Utility                                 & Named Entity Recognition                                          & Biological Concepts and Relations                       \\
\cite{Al-Aziz_etal_2022}             & 2021 & Privacy-preserving                    & Neural Network                    & DP-GPT                                                           & Publicly Available (MIMIC-III)                                & English           & NLL, BLEU, Jaccard, AdvSuc, DTP                                                                        & Similarity, Structure, Privacy, Test on Utility & Disease Classification                                            & Discharge Summaries                                     \\
\cite{RobertETAL2021}                & 2021 & Augmentation                          & Neural Network                    & Transformers                                                     & Online source                                                 & English           & Pattern Match, Classification Accuracy                                                                 & Structure, Test on Utility                      & Diagnose Prediction, Named Entity Recognition                     & History of Present Illness (HPI)                        \\
\cite{NingsihETAL2022}               & 2022 & Augmentation                          & Knowledge Source                  & Gazetteers                                                       & Online source                                                 & Indonesian        & Classification Accuracy                                                                                & Test on Utility                                 & Text Classification                                               & COVID News                                                    \\
\cite{FreiETAL2022}                  & 2022 & Annotation, Corpus Building           & Neural Network                    & GPT-neox                                                         & Prompting                                                     & German            & Pattern Match, Classification Accuracy                                                                 & Test on Utility                                 & Named Entity Recognition                                          & Drug and Medication                                     \\
\cite{TlachacETAL2022}               & 2022 & Augmentation                          & Neural Network                    & SeqGAN                                                           & Publicly Available                                            & English           & NLL, BLEU, Classification Accuracy                                                                     & Structure, Similarity, Test on Utility          & Phenotype Classification                                          & Doctor-Patient Conversations                            \\
\cite{HuangETAL2022}                 & 2022 & Assistive Writing                     & Neural Network                    & Bi-LSTM                                                          & Publicly Available                                            & Chinese           & Classification Accuracy                                                                                & Similarity                                      & Question Answering                                                & Medical Consultation                                    \\
\cite{XiaETAL2022}                   & 2022 & Assistive Writing                     & Neural Network                    & Transformers (Encoder-Decoder), SMedBERT                         & Publicly Available                                            & Chinese           & BLEU, Human Assessment                                                                                 & Similarity, Structure, Test on Utility          & Question Answering                                                & Medical Conversations                                   \\
\cite{WuETAL2022}                    & 2022 & Augmentation, Assistive Writing       & Neural Network                    & Transformers (T5)                                                & Publicly Available                                            & English           & ROUGE, BLEU, TER, Human Assessment                                                                     & Similarity, Structure, Test on Utility          & Report Generation                                                 & Patient Laboratory Test                                 \\
\cite{MoonETAL2022}                  & 2022 & Assistive Writing                     & Neural Network                    & VLP (CNN ResNet, BERT)                                           & Publicly Available (MIMIC-CXR)                                & English           & BLEU, Classification Accuracy                                                                          & Similarity, Test on Utility                     & Disease Classification, Report Generation, Question Answering     & Radiology                                               \\
\cite{WalonoskiETAL2022}             & 2022 & Corpus Building                       & Neural Network                    & GPT-2                                                            & Publicly Available (MIMIC-III)                                & English           & Human Assessment                                                                                       & Structure                                       & nan                                                               & Discharge Summaries (Cardiovascular)                    \\
\cite{Modersohn_etal_2022}           & 2022 & Corpus Building                       & Knowledge Source, Text Processing & UML, SpaCy                                                       & Online source                                                 & German            & Clustering                                                                                             & Structure                                       & nan                                                               & Clinical Practice Guidelines                            \\
\cite{Igarashi_Nihei_2022}           & 2022 & Augmentation                          & Text Processing                   & EDA                                                              & Manual Collection and Curation                                & Japanese          & Classification Accuracy                                                                                & Test on Utility                                 & Phenotype Classification                                          & Doctor-Patient Conversations                            \\
\cite{Ishikawa_etal_2022}            & 2022 & Augmentation                          & Neural Network                    & word2vec, Cosine Similarity                                      & Private EHR/EMR                                               & English           & Classification Accuracy                                                                                & Test on Utility                                 & Adverse Drug Reaction/Event Extraction                            & Drug and Medication                                     \\
\cite{Zhou_etal_2022_DataSifterText} & 2022 & Privacy-preserving                    & Knowledge Source, Text Processing & UML, RAKE                                                        & Publicly Available (MIMIC-III)                                & English           & Clustering, BLEU, TFIDF, Human Assessment (Turing Test)                                                & Similarity, Structure, Privacy                  & nan                                                               & Discharge Summaries                                     \\
\cite{FengETAL2022}                  & 2022 & Corpus Building                       & Neural Network                    & Transformers (T5 and BART)                                       & Manual Collection and Curation                                & English           & BLEU, ROUGE-L, CIDEr, METEOR, PPL, SPICE, BERTscore, Human Assessment                                  & Similarity, Structure                           & -                                                                 & History of Present Illness (HPI)                        \\
\cite{BelkadiETAL2023}               & 2023 & Augmentation                          & Neural Network                    & Transformers (T5, LT3)                                           & Publicly Available (MIMIC-III)                                & English           & BLEU, ROUGE, Jaccard, BERTscore                                                                        & Similarity                                      & Named Entity Recognition                                          & Drug and Medication                                     \\
\cite{YimETAL2023}                   & 2023 & Corpus Building                       & Knowledge Source, Neural Network  & UML, BioBART, LED                                                & Manual Collection and Curation                                & English           & ROUGE, BERTscore, BLEURT, MedCon                                                                       & Similarity                                      & -                                                                 & Doctor-Patient Conversations                            \\
\cite{BarretoETAL2023}               & 2023 & Augmentation                          & Neural Network                    & Transformers, CNN-ResNet                                         & Publicly Available                                            & English           & BLEU, Classification Accuracy                                                                          & Test on Utility                                 & Diagnose Prediction                                               & Radiology                                               \\
\cite{NguyenETAL2023}                & 2023 & Assistive Writing                     & Neural Network, Text Processing   & BART, distilBERT, Template-based                                 & Publicly Available                                            & English           & ROUGE                                                                                                  & Similarity                                      & Question Answering, Text Summarization                            & Doctor-Patient Conversations                            \\
\cite{KarakETAL2023}                 & 2023 & Assistive Writing                     & Neural Network                    & GPT-2                                                            & Publicly Available                                            & English           & BLEU, ROUGE                                                                                            & Similarity                                      & Question Answering                                                & Drug and Medication                                     \\
\cite{AggarwalETAL2023}              & 2023 & Augmentation                          & Neural Network                    & GPT-2                                                            & Publicly Available                                            & English           & ROUGE-L, PPL                                                                                           & Similarity, Structure                           & Named Entity Recognition                                          & Biological Concepts and Relations                       \\
\cite{MeyerETAL2023}                 & 2023 & Assistive Writing                     & Neural Network                    & Seq2Seq (RNN Encoder and RNN Decoder) Point Generator Network    & Online source                                                 & English           & ROUGE, ROUGE-L, Cosine, Jaccard, TFIDF                                                                 & Similarity                                      & Text Summarisation                                                & Drug and Medication                                     \\
\cite{BenAbachaETAL2023}             & 2023 & Corpus Building, Augmentation         & Neural Network                    & BART                                                             & Private EHR/EMR                                               & English           & ROUGE, BLEURT, BERTScore, Human Assessment                                                             & Similarity, Structure                           & Text Summarisation                                                & Doctor-Patient Conversations                            \\
\cite{TangETAL2023}                  & 2023 & Augmentation, Annotation, Usefulness  & Neural Network                    & ChatGPT                                                          & Prompting                                                     & English           & Classification Accuracy, Human Assessment, Pattern Match                                               & Similarity, Structure, Test on Utility          & Named Entity Recognition, Relation Extraction                     & Biological Concepts and Relations                       \\
\cite{SchlegelETAL2023}              & 2023 & Augmentation                          & Neural Network, Knowledge Source  & ChatGPT-3.5-turbo, UML                                           & Publicly Available (MIMIC-III)                                & English           & ROUGE, BERTscore, BLEURT                                                                               & Similarity, Test on Utility                     & Text Summarisation                                                & Medical Conversations                                   \\
\cite{HiebelETAL2023}                & 2023 & Usefulness                            & Neural Network                    & GPT-2                                                            & Publicly Available                                            & French            & BLEU, Pattern Match, Human Assessment                                                                  & Similarity, Structure, Test on Utility          & Named Entity Recognition                                          & Biological Concepts and Relations                       \\
\cite{PengETAL2023}                  & 2023 & Usefulness                            & Neural Network                    & GPT-3                                                            & Private EHR/EMR                                               & English           & Human Assessment (Turing Test), Classification Accuracy                                                & Structure, Test on Utility                      & De-Identification, Relation Extraction, Question Answering        & History of Present Illness (HPI)                        \\
\cite{IrfanETAL2023}                 & 2023 & Augmentation                          & Neural Network                    & GPT (ChatGPT, Bard)                                              & Publicly Available                                            & English           & BERTScore, Classification Accuracy                                                                     & Similarity, Test on Utility                     & Phenotype Classification                                          & Doctor-Patient Conversations (Mental Health)            \\
\cite{LiETAL2023}                    & 2023 & Augmentation, Annotation              & Neural Network                    & GPT-4                                                            & Manual Collection and Curation, Public Available. (MIMIC-III) & English           & Human Assessment (Pattern Match), Classification Accuracy                                              & Structure, Test on Utility                      & Phenotype Classification                                          & History of Present Illness (HPI)                        \\
\cite{XuETAL2024}                    & 2023 & Augmentation, Privacy-preserving      & Neural Network                    & ChatGPT, PubMedBERT                                              & Prompting                                                     & English           & Classification Accuracy                                                                                & Test on Utility                                 & Relation Extraction, Named Entity Recognition, Question Answering & Clinical Transcripts                                    \\
\cite{ReynoldsETAL2024}              & 2024 & Assistive Writing                     & Neural Network                    & ChatGPT                                                          & Prompting                                                     & English           & Human Assessment                                                                                       & Structure, Test on Utility                      & Question Answering                                                & Medical Consultation                                    \\
\cite{JiETAL2024}                    & 2024 & Assistive Writing                     & Neural Network                    & ClinicalBLIP                                                     & Publicly Available (IU X-RAY, MIMIC-CXR)                      & English           & ROUGE, METEOR                                                                                          & Simlarity                                       & Report Generation                                                 & X-Ray                                                   \\
\cite{LatifETAL2024}                 & 2024 & Augmentation                          & Neural Network, Text Processing   & ChatGPT, BART, T5, EDA                                           & Publicly Available                                            & English           & ROUGE, CIDEr, METEOR, BERTscore                                                                        & Similarity                                      & Diagnose Prediction, Disease Classification                       & Population, Intervention, Comparison, and Outcome (PICO)\\
\cite{CloughETAL2024}                & 2024 & Usefulness                            & Neural Network                    & ChatGPT                                                          & Prompting                                                     & English           & Human Assessment                                                                                       & Structure Comparison (Human vs. Synthetic text) & Report Generation                                                 & Discharge Summaries                                     \\
\cite{QUETAL2023}                    & 2023 & Assistive Writing                     & Neural Network                    & GPT                                                              & Online source                                                 & Chinese           & PPL                                                                                                    & Structure, Test on Utility                      & Question Answering                                                & Doctor-Patient Conversations                            \\
\cite{BiswasETAL2024}                & 2024 & Assistive Writing                     & Neural Network                    & VAE, GAN (LSTM Generator, CNN Discriminator)                     & Publicly Available (MIMIC-III)                                & English           & BLEU, PPL, WER, Human Assessment                                                                       & Structure, Similarity                           & -                                                                 & Clinical Transcripts                                    \\
\cite{ZhaoETAL2024}                  & 2024 & Assistive Writing                     & Neural Network                    & BART                                                             & Publicly Available                                            & English           & ROUGE-L, BERTscore                                                                                     & Similarity                                      & Report Generation                                                 & Patient Laboratory Test                                 \\
\cite{DelmasETAL2024}                & 2024 & Augmentation                          & Neural Network                    & BioGPT                                                           & Publicly Available                                            & English           & BLEU, Classification Accuracy                                                                          & Similarity, Test on Utility                     & Relation Extraction                                               & Biological Concepts and Relations                       \\
\cite{ZecevicETAL2024}               & 2024 & Assistive Writing, Privacy-preserving & Neural Network                    & DP-GPT, BioGPT                                                   & Private EHR/EMR                                               & English           & Human Assessment, ROUGE-L, Privacy Budget                                                              & Similarity, Structure, Privacy                  & Report Generation                                                 & Endoscopy Reports                                       \\
\cite{SinghETAL2024}                 & 2024 & Augmentation                          & Neural Network                    & Mistral, Llama, Gemma                                            & Publicly Available                                            & English           & Jaccard, BERTscore, Human Assessment                                                                   & Similarity, Structure                           & De-Identification                                                 & Discharge Summaries                                     \\
\cite{SanjeevETAL2024}               & 2024 & Assistive Writing                     & Neural Network                    & Transformers                                                     & Publicly Available (MIMIC-CXR)                                & English           & ROUGE, BLEU, METEOR, Human Assessment                                                                  & Similarity, Structure                           & Report Generation                                                 & Radiology                                               \\
\cite{BaiETAL2024}                   & 2024 & Assistive Writing                     & Neural Network                    & Transformers (CLIP)                                              & Publicly Available                                            & English           & BLEU, ROUGE, METEOR, BERTScore                                                                         & Similarity, Test on Utility                     & Report Generation, Question Answering                             & Radiology                                               \\
\cite{MeoniETAL2024}                 & 2024 & Privacy-preserving                    & Neural Network, Knowledge Source  & GPT (Mistral), RL, UML                                           & Publicly Available (MIMIC-III)                                & English           & SemScore                                                                                               & Similarity, Privacy                             & -                                                                 & Discharge Summaries                                     \\
\cite{LiuETAL2024}                   & 2024 & Assistive Writing                     & Neural Network                    & Transformers (PLM), GPT (Galactica)                              & Publicly Available                                            & English           & BLEU, ROUGE, ROUGE-L, METEOR                                                                           & Similarity                                      & Question Answering                                                & Biological Concepts and Relations                       \\
\cite{HughesETAL2024}                & 2024 & Usefulness                            & Neural Network                    & Transformers (T5)                                                & Online source                                                 & English           & Classification Accuracy, Human Assessment                                                              & Structure, Test on Utility                      & Named Entity Recognition                                          & Population, Intervention, Comparison, and Outcome (PICO)\\
\cite{ZhanETAL2024}                  & 2024 & Assistive Writing                     & Neural Network                    & LDM (VAE, CNN ResNet)                                            & Publicly Available (IU X-RAY, MIMIC-CXR)                      & English           & BLEU, ROUGE-L                                                                                          & Similarity, Test on Utility                     & Report Generation                                                 & X-Ray, MRI, CT-scan                                     \\
\cite{CalvoLorenzoETAL2024}          & 2024 & Usefulness                            & Neural Network                    & ChatGPT-3.5-turbo                                                & Prompting                                                     & English           & Human Assessment                                                                                       & Structure                                       & -                                                                 & History of Present Illness (HPI)                        \\
\cite{LundETAL2024}                  & 2024 & Annotation, Corpus Building           & Neural Network                    & GPT-4                                                            & Private EHR/EMR                                               & Norwegian         & Human Assessment, Classification Accuracy                                                              & Similarity, Structure, Test on Utility          & De-Identification                                                 & Discharge Summaries                                     \\
\cite{ParresETAL2024}                & 2024 & Assistive Writing                     & Neural Network                    & Transformers (VED)                                               & Publicly Available (MIMIC-CXR)                                & English           & BLEU, ROUGE-L, F1cXb                                                                                   & Similarity, Test on Utility                     & Report Generation                                                 & Radiology                                               \\
\cite{LitakeETAL2024}                & 2024 & Usefulness                            & Neural Network                    & ChatGPT-3.5                                                      & Publicly Available (MIMIC-III), Prompting                     & English           & Classification Accuracy                                                                                & Test on Utility                                 & Disease Classification                                            & Discharge Summaries                                     \\
\cite{OhETAL2024}                    & 2024 & Usefulness                            & Neural Network                    & ChatGPT-3.5-Turbo, BioGPT, GPT-2, distilGPT2, CerebroGPT         & Private EHR/EMR                                               & English           & BLEU, ROUGE, Cosine, TF-IDF, Human Assessment                                                          & Similarity, Structure, Test on Utility          & Report Generation                                                 & Radiology (Cerebrovascular)                             \\
\cite{ALMutairiETAL2024}             & 2024 & Corpus Building                       & Neural Network                    & Claude-3-Opus, GPT-4                                             & Publicly Available                                            & English, Arabic   & ROUGE-L, BERTScore, Human Assessment                                                                   & Similarity, Structure                           & -                                                                 & Doctor-Patient Conversations                            \\
\cite{MitraETAL2024}                 & 2024 & Corpus Building                       & Neural Network                    & GPT-4                                                            & Manual Collection and Curation, Prompting                     & English           & ROUGE-L, Human Assessment, Classification Accuracy                                                     & Similarity, Structure, Test on Utility          & Named Entity Recognition                                          & History of Present Illness (HPI)                        \\
\cite{KotschenreutherETAL2024}       & 2024 & Corpus Building                       & Neural Network                    & GPT (Llama)                                                      & Publicly Available                                            & English           & Human Assessment                                                                                       & Structure, Test on Utility                      & Question Answering                                                & Discharge Summaries                                     \\
\cite{SunETAL2023}                   & 2025 & Assistive Writing                     & Neural Network                    & Transformers (BART)                                              & Publicly Available                                            & English           & ROUGE, BLEU                                                                                            & Similarity                                      & Question Answering                                                & Doctor-Patient Conversations                            \\
\cite{BiniciETAL2024}                & 2025 & Augmentation                          & Neural Network                    & Llama, Mistral, Gemma                                            & Publicly Available                                            & English           & BLEU, ROUGE, ROUGE-L, WER, BERTScore, Human Assessment                                                 & Similarity, Structure                           & -                                                                 & Doctor-Patient Conversations                            \\
\cite{ScrogginsETAL2024}             & 2024 & Augmentation                          & Neural Network                    & GPT-4                                                            & Prompting                                                     & English           & Classification Accuracy                                                                                & Test on Utility                                 & Diagnose Prediction                                               & Medical Conversations                                   \\
\cite{LorgeETAL2024}                 & 2025 & Augmentation, Annotation              & Neural Network                    & ChatGPT-3.5                                                      & Private EHR/EMR, Prompting                                    & English           & Classification Accuracy, Pattern Match                                                                 & Structure, Test on Utility                      & Phenotype Classification                                          & History of Present Illness (HPI)                        \\
\hline
\end{longtblr}
\end{landscape}

\end{document}